\documentclass[journal,10pt]{IEEEtran}

\usepackage{amsmath,amssymb,amsfonts}
\usepackage{tikz}
\usepackage{pgfplots}
\usepackage{textcomp}
\usepackage{float}
\usepackage{algorithm2e}
\usepackage{comment}
\RestyleAlgo{ruled}
\usepackage{subcaption}
\usepackage{graphicx}
\usepackage{textcomp}
\usepackage{xcolor,flushend}
\IEEEoverridecommandlockouts

\newcommand{\floor}[1]{\left\lfloor #1 \right\rfloor}
\newcommand{\ceil}[1]{\left\lceil #1 \right\rceil}

\usepackage{etoolbox}
\makeatletter
\patchcmd{\@IEEEbiography}
  {\vskip 6pt\relax\nobreak\bigskip\nobreak}
  {\vskip 0pt\relax} 
  {}{}
\patchcmd{\@IEEEbiography}
  {\bigskip}
  {\vskip 0pt}
  {}{}
\AtBeginEnvironment{IEEEbiography}{\parskip=0pt\parindent=0pt}
\makeatother

\begin{document}
\title{Evasive Active Hypothesis Testing with Deep Neuroevolution: The Single- and Multi-Agent Cases}
\author{
    George Stamatelis,~\IEEEmembership{Student~Member,~IEEE}, Angelos-Nikolaos Kanatas,~\IEEEmembership{Student~Member,~IEEE}, \\ Ioannis Asprogerakas, and George C. Alexandropoulos,~\IEEEmembership{Senior~Member,~IEEE} 
   \thanks{Part of this work has been presented in the IEEE 
International Conference on Communications (ICC), Denver, CO, USA, June 2024~\cite{eahtConf}.}
    \thanks{G. Stamatelis and G. C. Alexandropoulos are with the Department of Informatics and Telecommunications, National and Kapodistrian University of Athens, Panepistimiopolis Ilissia, 15784 Athens, Greece. G. C. Alexandropoulos is also with the Department of Electrical and Computer Engineering, University of Illinois Chicago, IL 60601, USA 
    (e-mails: \{georgestamat, alexandg\}@di.uoa.gr).}
    \thanks{A. -N. Kanatas and I. Asprogerakas are with the School of Electrical and Computer Engineering, National Technical University of Athens, Zografou Campus, 15780 Athens, Greece
    (e-mails: \{el19169, el18942\}@mail.ntua.gr).}
\thanks{The research work has been supported by the Hellenic Foundation for Research and Innovation (HFRI) under the 5th Call for HFRI PhD Fellowships (Fellowship Number: 21080), and the Smart Networks and Services Joint Undertaking project 6G-DISAC under the European Union's Horizon Europe research and innovation programme under Grant Agreement No 101139130. 
}
}     

\maketitle
\begin{abstract}
Active hypothesis testing is a thoroughly studied problem that finds numerous applications in wireless communications and sensor networks. In this paper, we focus on one centralized and one decentralized problem of active hypothesis testing in the presence of an eavesdropper. For the centralized problem including a single legitimate agent, we present a new framework based on deep NeuroEvolution (NE), whereas, for the decentralized problem, we develop a novel NE-based method for solving collaborative multi-agent tasks, which, interestingly, maintains all computational benefits of our single-agent NE-based scheme. To further reduce the computational complexity of the latter scheme, a novel multi-agent joint NE and pruning framework is also designed. The superiority of the proposed NE-based evasive active hypothesis testing schemes over conventional active hypothesis testing policies, as well as learning-based methods, is validated through extensive numerical investigations in an example use case of anomaly detection over wireless sensor networks. It is demonstrated that the proposed joint optimization and pruning framework achieves nearly identical performance with its unpruned counterpart, while removing a very large percentage of redundant deep neural network weights.
\end{abstract}


\section{Introduction}
Active Hypothesis Testing (AHT) refers to the family of problems where one legitimate agent or decision  maker, or a group of collaborating agents or decision makers, adaptively select(s) sensing actions and collect(s) observations in order to infer the underlying true hypothesis in a fast and reliable manner~\cite{ChernoffHT,controlledSensForMH}. AHT and related active sensing problems, such as active parameter estimation~\cite{activeSamplingMultiSource} and active change point detection~\cite{banditChangePoint}, find numerous applications in wireless communications, including anomaly detection over sensor networks~\cite{DRL-AHT-AD-journal,josephScalableAD}, strong or weak radar models for target detection~\cite{chernoff-radar}, camera object detection \cite{cameraObjectDetection}, cyber-intrusion detection, target search, and adaptive beamforming~\cite{activeLearningMMWave}, as well as, very recently, localization \cite{activeSensingRisLocalisation} and channel estimation~\cite{ActiveSensingComLearn} enabled by reconfigurable intelligent surfaces. 

The recent rise of distributed and edge machine learning approaches~\cite{ASH2022,edge_ml_survey2}, as well as Internet-of-Things (IoT) applications \cite{iot_survey}, is urging the development of efficient mechanisms for large-scale covert data collection. It has been shown in~\cite{IoTexposure} that, even in encrypted IoT applications, eavesdroppers can accurately estimate sensitive information just by observing device interactions alone. The main focus of this paper is on decentralized collaboration mechanisms for active sensing that do not reveal information to third parties.
\color{black}
\subsection{Background}

\paragraph{Deep Reinforcement Learning (DRL)} Reinforcement Learning (RL) and especially DRL, which leverages the representation capabilities of Deep Neural Networks (DNNs), has emerged as a very powerful tool for complex decision making in modern wireless communication systems. The seminal paper on Deep Q Networks (DQN)~\cite{mnih2015humanlevel} for video games and subsequent works on policy gradient methods, e.g. \cite{schulman2017ppo,asynchronousDRL} for robotics, laid the foundation for profound resource allocation performance in a wide range of communication systems. Although DRL can be used to solve traditional Markov Decision Processes (MDPs), its success is mainly attributed to its capabilities to find very good, near-optimal policies for Partially Observable MDPs (POMDPS), which are known to be NP-hard problems~\cite{Papadimitriou1987TheCO}.

\color{black}
\paragraph{Multi-Agent Systems and DRL}
When it comes to collaborative multi-agent MDPs and POMDPs, state-of-the-art DRL approaches \cite{wong2022deepMARLSurvey,multiagentRLTheory} are based on the idea of centralized learning and decentralized execution (CLDE) \cite{CLDE-Rehersal}. During training, the agents are provided with additional information that enhances the training process. The agent, however, must not depend on that information during deployment/testing. Most popular CLDE algorithms for tasks with heterogeneous agents are based on the \textcolor{black}{Multi-Agent Deep Deterministic Policy Gradients} (MADDPG) actor critic algorithm \cite{multiACMixed}, where each agent is equipped with an individual actor and there is a global critic DNN. \textcolor{black}{This algorithm finds numerous applications in wireless communication systems, including cognitive radio \cite{MADDPGSpectrum}, power control \cite{MADDPGPowerControl}, and edge caching \cite{MADDPGCaching}}. Extensions of MADDPG based on Proximal Policy Optimization (PPO)~\cite{schulman2017ppo} have been successfully applied to AHT \cite{decHTKohen,stamatelisColab}. 
Furthermore, federated extensions of MADDPG have been recently discussed in \cite{fmaddpg1,fmaddgp2}.
\paragraph{DNN Pruning}
There is lately an increased demand for deploying pre-trained DNNs on lighter devices with memory and/or power constraints~\cite{stateNetPrun,HanPrun}, such as mobile phones, lightweight sensors, and various IoT. However, running large, over-parameterized DNNs on such devices is often impossible. DNN pruning algorithms~\cite{HanPrun,lecunPrune} remove unnecessary connections and/or neurons in order to get smaller neural networks with similar performance. Such algorithms are based on the lottery ticket hypothesis which states~\cite{LTH}: ``Random dense feed-forward
DNNs contain \textit{winning tickets}, i.e., smaller subnetworks that can achieve almost identical performance to the initial network when trained alone.'' To this end, pruning has been successfully applied to single-agent DLR problems~\cite{RLpruneGraeser,rlPrunStone} and, more recently, to multi-agent DRL settings~\cite{marlPrune}. 

\paragraph{DRL for AHT}
The problem of binary AHT was first studied by Chernoff in his pioneering work on sequential design of experiments~\cite{ChernoffHT}. This work proposed an asymptotically optimal heuristic, known as the Chernoff test, which remains popular even today. The Chernoff test was latter extended in the multi-hypothesis setting~\cite{controlledSensForMH}.
In~\cite{Naghshvar_2013, stamatelis2023active, ac-ad-noCost, KartrikDQN}, AHT was modeled  as a POMDP. The authors in~\cite{Naghshvar_2013} presented bounds based on dynamic programming, whereas~\cite{stamatelis2023active,ac-ad-noCost,KartrikDQN} showcased the superiority of DRL strategies over conventional AHT heuristics. The recurrent DRL algorithm in~\cite{stamatelis2023active} was shown to compete with classical model-based strategies without having knowledge of the environment dynamics. More complex AHT-based anomaly detection problems with sampling costs have recently attracted a lot of attention, e.g.,~\cite{AC-AnomDetectProbCost,sacAD,monitoringAC}, and appropriate deep learning and DRL strategies that balance detection objectives with cost management were proposed. Collaborative multi-agent DRL for AHT was studied in~\cite{decHTKohen,josephScalableAD,stamatelisColab}. Specifically, the authors in~\cite{stamatelisColab} discussed how sampling cost constraints can be managed in a multi-agent environment using Lagrange multipliers. Very recently in~\cite{RLADVAD,controlledSensingCorrupted,aaht}, AHT in the presence of adversaries that target to corrupt the observations of legitimate agents was studied. The first two works assumed no adaptive decision making from the adversary's side and, in fact, in \cite{controlledSensingCorrupted} the agent terminated when an adversary was detected. The last work focused on the case of adaptive and intelligent legitimate as well as adversarial agents with different information structures. 

\paragraph{NeuroEvolution(NE)}
The consideration of NE schemes for solving MDPs and POMDPs is an old idea, dating back, for example, to \cite{moriarty:mlj96,YaoNEProceedinga}, which has been left largely undeployed, mainly due to the recent impressive success of DRL approaches. However, in the past few years, it has been experimentally shown that, even simple NE schemes, can rival back-propagation algorithms, such as deep Q-learning and policy gradients, outperforming DRL approaches in various single-agent POMDPs \cite{salimans2017evolution,benchmaringBasics,MBACNNJournal}. Surprisingly, even very old NE methods can compete with popular state-of-the-art DRL algorithms, as shown in \cite{benchmaringBasics}. The main benefits of NE over DRL are summarized as follows: 
\begin{itemize}
    \item NE is easier to implement (replay buffers, advantage estimation, etc., are not needed) and to parallelize over multiple Central Processing Units (CPUs). Only scalar numbers indicating the fitness of an individual need to be shared between collaboratively computing nodes.
    \item Reward reshaping and exploration techniques are not required in NE schemes. It is well known in RL and DRL literature that training algorithms with very sparse reward signals rarely produce good performance, and  designing appropriate rewards can be a time-consuming trial and error task. On the other hand, NE only needs to specify a fitness function.  This benefit comes extremely handy in decision-making  problems that have multiple constraints besides their core objective, such as the ones studied in this paper for secure active sensing. 
    \item DRL schemes face instability problems, which are associated with back-propagation through time. This issue is totally absent in NE-based approaches.
\end{itemize}

The core idea of NE is to directly search the space of policy DNNs via nature-inspired algorithms; note that, in NE, critic DNNs are not considered. In particular, each chromosome of an individual represents some parameters of a policy DNN \cite{salimans2017evolution,cosyne}. It is noted that, due to the large number of parameters in a DNN, it is typically infeasible to construct individuals representing all of the DNN parameters. To this end, techniques that take advantage of the DNN's structure in order to construct smaller individuals are usually devised. Particularly, a generation of individuals is initialized randomly. Each individual is then evaluated, and its fitness function is stored. The individuals with the highest fitness function are selected for mating. During mating, the parameters of two or more individuals are merged by various methods (e.g., crossover operation). The new individuals then replace the "weaker" individuals of the population. This procedure is repeated for multiple generations. It is noted that further genetic operations~\cite{holland}, such as mutation, can be utilized to increase exploration.

However, despite the recent impressive results of NE schemes for single-agent problems, to the best of our knowledge, there exist no works elaborating on how to extend them to multi-agent collaborative problems, which is the focus of this research work. Furthermore, this paper constitutes the first attempt at applying pruning methods on evolved policy DNNs.

\paragraph{Private Hypothesis Testing}
Due to the growing concerns for data privacy, many works studied privacy in passive hypothesis testing problems, where there is no control over the sensing actions. For example, differentially private hypothesis testing was studied in~\cite{DifferntPrivateHT}, whereas~\cite{stateEstimEve} elaborated on how to perform remote estimation of the system state through sensor data while impairing the filtering ability of eavesdroppers. Secure distributed hypothesis testing was studied in~\cite{secureDistrHT}. A closely related problem is the active privacy utility trade-off in data sharing~\cite{activePUT,PrivacyAwarDatasharing}. In the problem studied in \cite{activePUT}, there are two independent discrete variables $S$ and $U$, and the observations generated depend on both. The DRL agent adaptively selects data release mechanisms and outputs observations to a service provider. The goal is to assist the service provider in determining the value of $U$, while keeping $S$ hidden. In contrast, the authors in~\cite{PrivacyAwarDatasharing} investigated real-time data sharing methods for a Markov chain $X_t$, where the objective is to ``hide'' the true value of $X_t$ at each time step, while ensuring that the distortion between the shared observations $Y_t$ and the actual $X_t$ remains below a pre-defined threshold. These formulations differ from our work because our agent tries to both infer and hide the same variable. Besides that, the latter data sharing frameworks are only limited to a single-agent (centralized) scenario.

The problem of single-agent Evasive Active Hypothesis Testing (EAHT), where a passive eavesdropper (Eve) collects noisy estimates of the legit observations and tries to infer the underlying hypothesis, was studied in~\cite{evasive-active}, focusing however explicitly on the asymptotic case. In that work, the authors formulated single-agent EAHT as a constrained optimization problem including the legitimate agent's and Eve's error exponent. However, near-optimal or optimal action selection policies were not presented. In this paper, motivated by the lack of explicit policies for EAHT, we present novel single- and multi-agent EAHT approaches for wireless sensor networks, which are both based on a deep NE framework. The contributions of this paper are summarized as follows:
\vspace{-0.1cm}
\begin{enumerate}
    \item We formulate the single-agent EAHT problem studied in~\cite{evasive-active} as a constrained POMDP and present a NE-based method for solving it. Our numerical investigations showcase that our method satisfies the privacy constraint, while achieving similar accuracy to popular AHT methods that ignore the existence of any adversary. 
    \item A novel formulation of the decentralized multi-agent EAHT problem is presented, where a group of agents tries to infer the underlying hypothesis, while keeping it hidden from a passive eavesdropper.
    \item We present a novel approach for solving decentralized POMDPs via deep NE, and apply it to the decentralized EAHT problem at hand. The proposed scheme is numerically compared with state-of-the-art multi-agent DRL algorithms. It is demonstrated that our NE-based method outperforms existing algorithms, while maintaining all computational benefits of our single-agent NE scheme.
    \item A novel multi-agent joint NE and pruning scheme is devised, which is shown experimentally to achieve almost identical performance to the unpruned agents, despite removing over $90\%$ of the DNN's weights.
\end{enumerate}

This paper extends its recent conference version~\cite{eahtConf} by including a novel multi-agent joint NE and pruning scheme, as well as more thorough experiments including more benchmark schemes, a second synthetic sensor model, additional wireless applications, and new experiments against sophisticated, learning-based eavesdroppers.

The remainder of this paper is organized as follows. Section~II introduces the single-agent (centralized) EAHT problem and its multi-agent (decentralized) extension. Section~III presents our NE-based solution methods, and Section~IV includes our extensive experimental results demonstrating the superiority of the proposed EAHT schemes over various benchmarks. The paper is concluded in Section~V. 

\subsection{Notations}
Throughout this paper,  calligraphic letters, e.g. $\mathcal{X}$, are reserved for sets. Bold lower-case and upper-case letters denote vectors and matrices, respectively, e.g., $\boldsymbol{\theta}$ and $\boldsymbol{\Theta}$. Notation $[\boldsymbol{\Theta}]_{l,m}$ denotes the element on the  $l$-th row and $m$-th collumn of the matrix $\boldsymbol{\Theta}$. Unless stated otherwise, the  letter $t$ is reserved for time indices. Finally, $E[\cdot]$ represents the expectation operator, while $\hat{E}[\cdot]$ denotes the sample average.

\section{EAHT Problem Formulations}\label{sec:problem_form}
Consider a security analyst monitoring a corporate network for signs of intrusion. The analyst (or an automated detection system) can deploy various tests, such as port scans, access log queries, or anomaly detection filters, to identify whether the system is under attack and, if so, determine the type and location of the threat. However, an adversary might be monitoring these tests as well to identify weakened components of the network with the goal to launch more tailored attacks. As another application, consider search and rescue missions, where a team of autonomous drones collaborates to locate survivors in a disaster zone (e.g., an earthquake-hit city), while avoiding detection by hostile actors (e.g., armed groups or adversarial drones). The swarm shares partial observations (e.g., thermal signatures and/or structural damage) to rapidly narrow down survivor locations, but carefully controls communication timing and searching actions to prevent eavesdroppers from inferring their progress. Some drones may even emit decoy signals or take deceptive patrol routes to distort the adversary’s belief distribution. While timely detection of survivors is the main goal, the swarm may also desire to ensure that the eavesdropper(s) cannot assign high confidence to a single hypothesis (e.g., that survivors exist in a specific building).

In this paper, we study methods to collect informative data in order to accurately classify the underlying state of a system, while keeping it hidden from eavesdropping third parties. Two EAHT problems are introduced in this section.
A centralized one with a single agent, and a decentralized one with a group of agents having access to different sensing action sets. In the latter problem, the agents exchange information with each other and each one separately infers the hypothesis~\cite{decHTKohen}. 

\subsection{Centralized Problem}
Let $\mathcal{X}\triangleq\{0,1,2,\ldots,|\mathcal{X}|-1\}$ be a finite set of hypotheses, while the true hypothesis $x$ is unknown.  A legitimate agent has access to a finite set of sensing actions $\mathcal{A}$, and at each time instance $t$, it selects an action $a_t\in\mathcal{A}$. In response to this action, the agent collects a noisy observation $y_t$. In parallel, an eavesdropper (Eve) being present in the system receives another noisy observation $z_t$. 
The conditional probability of $y_t$ given $x$ and $a_t$ is denoted as $P[y_t|a_t,x]$, whereas the respective conditional probability of $z_t$ is $Q[z_t|a_t,x]$.

We assume that the prior over all hypotheses $\pi_0(\mathcal{X})$ and the  distributions $P[\cdot]$ and $Q[\cdot]$ are either known a priori \cite{Naghshvar_2013,evasive-active}, or can be reliably estimated from a large dataset. However,  we will also experimentally verify the effectiveness of the proposed strategies in environments where the estimated probability kernels are incorrect approximations of the true dynamics. 
\color{black}

The legitimate agent maintains an $|\mathcal{X}|$-dimensional belief vector $\pi^L_t(\mathcal{X})$ over all possible hypotheses $x \in \mathcal{X}$ at time instant $t$, and Eve does the same via the belief vector $\pi^E_t(\mathcal{X})$. 
Each entry $\pi_t^L(x)$ is the posterior probability on the hypothesis $x$, given the sequence of action and observations up to time $t$.
For the former, given an action observation pair $(a_t,y_t)$, the legit belief on each hypothesis $x$ is updated as follows~\cite{KartrikDQN}: 
\begin{equation}
\label{legBelUP}
    \pi^L_t(x)=\frac{\pi^L_{t-1}(x)P[y_t|a_t,x]}{\sum_{x' \in \mathcal{X}}\pi^L_{t-1}(x')P[y_t|a_t,x']}.
\end{equation}
Similarly, Eve's belief given a pair $(a_t,z_t)$ can be updated as: 
\begin{equation}
\label{eaveBelUP}
    \pi^E_t(x)=\frac{\pi^E_{t-1}(x)Q[z_t|a_t,x]}{\sum_{x' \in \mathcal{X}}\pi^E_{t-1}(x')Q[z_t|a_t,x']}.
\end{equation}

By assuming that both agents deploy the optimal Maximum A Posteriori (MAP) decoding~\cite{evasive-active,aaht} the error probabilities at each time instant $t$ can be expressed as follows: 
\begin{align}
    \gamma^L_t = 1-\max_{x \in \mathcal{X}}\pi^L_t(x),\\
    \gamma^E_t= 1-\max_{x \in \mathcal{X}}\pi^E_t(x).
\end{align}
\color{black}
We also assume that the legitimate agent controls the stopping time\footnote{In the context of AHT, the stopping time metric refers to the number of sensing actions performed before the final inference decision~\cite{josephScalableAD,stamatelis2023active,sacAD}.} $\tau$. To this end, once the episode terminates, both the agent and Eve guess the underlying hypothesis according to the maximum a posteriori decoding rule.

The goal of the legitimate agent is to reliably estimate the true hypothesis as quickly as possible while keeping Eve's error probability above a certain application/agent-defined threshold.
Let $g_t\triangleq g(a_t|\pi^L_t(\mathcal{X}))$ represent the policy of the agent at each time instance $t$. The policy is a probabilistic mapping from belief vectors to the action set. Hence, the total policy of the legitimate agent for a sensing horizon of $\tau$ time slots is defined as follows:
\begin{equation}
\label{eq:policyDef}
    {\rm g}\triangleq(g_1,g_2 \cdots, g_\tau,\tau).
\end{equation}
By using user defined scalars $E$ and $L$, this problem can be formulated as a constrained POMDP problem as follows: 
\begin{align}
    \mathcal{OP}_1&:\nonumber\underset{{\rm g}}{\min} \,\, E[\tau]\\
    &\nonumber\quad\text{\text{s}.\text{t}.}\,\,\,\,\, \gamma^L_\tau \leq L,\,\,\gamma^E_t \geq E\,\,\,\,\forall t=1,2,\ldots,\tau.
\end{align}
%
The expectation in the latter objective is taken with respect to ${\rm g}$, $P[\cdot]$, $Q[\cdot]$, and $\pi_0(x)$. This indicates that the prior and probability kernels influence the belief updates in~(\ref{legBelUP}) and~(\ref{eaveBelUP}), which in turn influence future policies, beliefs, and decision rules. Note that, without the second constraint, $\mathcal{OP}_1$ is essentially an AHT problem.  \color{black}It is also noted that, in practice, the error probabilities  cannot be accurately revealed through a finite number of episodes, implying that there will always be a non-zero probability of leakage. To deal with this issue, we use sample averaging to simplify the constraints as follows:
\begin{equation}\label{eq:RelaxedConstraints}
\hat{E}[1-\max_{x \in \mathcal{X}} \pi_\tau^L(x)] \leq L, \quad 
    \hat{E}[1-\max_{x \in \mathcal{X}} \pi_t^E(x)] \geq E\,\,\forall t.
\end{equation}
\color{black}

Even without the instantaneous constrains, POMDPs are NP-hard \cite{Papadimitriou1987TheCO}, therefore, we do not expect to find exact solutions. In the next section, we will present near-optimal policies using deep policy optimization techniques.

\color{black}

\subsection{Decentralized Problem}
\label{sec:DedDef}
\begin{figure}
    \centering
    \includegraphics[width=\linewidth]{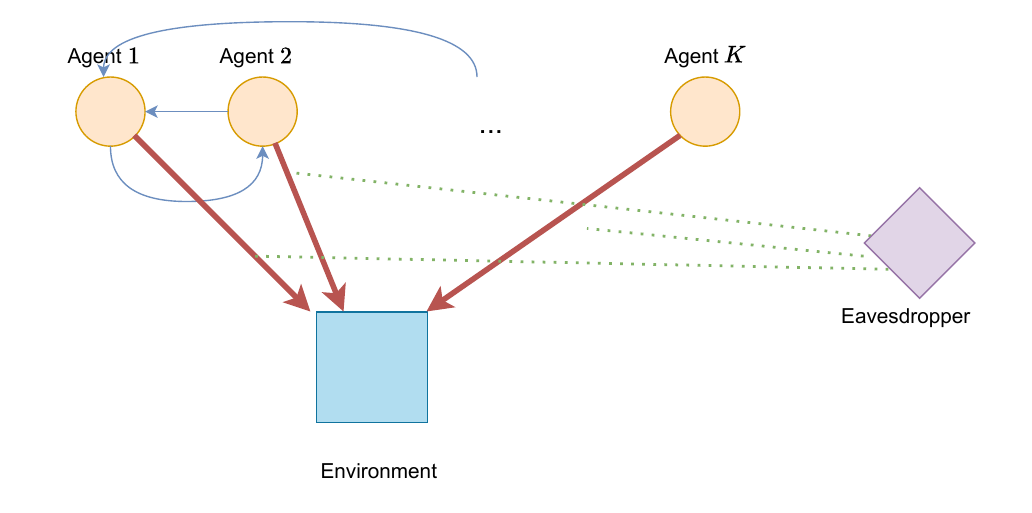}
    \caption{The decentralized EAHT problem under consideration. The agents monitor the environment through their sensing actions (thick red arrows) and, at the same time, they share information with each other (blue arrows). In their vicinity, there exists an eavesdropper monitoring their actions and collecting the corresponding observations through a noisy channel (dashed green lines).}
    \label{fig:decEAHT-setup}
\end{figure}

In the decentralized problem depicted in Fig. \ref{fig:decEAHT-setup}, a group of $K$ legitimate agents collaborates to infer the underlying hypothesis. We assume that each $k$-th agent, with $k=1,2,\ldots,K$, has access to a sensing action set $\mathcal{A}^k$. By probing the environment with an action $a^k_t$ at time slot $t$, the $k$-th agent receives a noisy observation $y_t^k$ with conditional distribution $P[y_t^k|a_t^k,x]$, while Eve observes the noisy quantity $z_t^k$ with conditional distribution $Q[z_t^k|a_t^k,x]$. It is assumed that both $y_t^k$ and $z_t^k$ do not depend on the actions of other than the $k$-th agent; such assumptions are common in the collaborative anomaly detection literature, e.g., \cite{josephScalableAD,decHTKohen}. This agent is assumed to also broadcast the action and observation tuple $(a_t^k,y_t^k)$ to a set $\mathcal{O}^k$ of neighboring legitimate agents, while receiving observations from another set $\mathcal{I}^k$ of agents.

Given a pair $({\rm a}_t,{\rm y}_t)$ of actions and observations, with ${\rm a}_t\triangleq(a_t^1,a_t^2,\ldots,a_t^K)$ and ${\rm y}_t\triangleq(y_t^1,y_t^2,\ldots,y_t^K)$, each $k$-th legitimate agent updates its belief according to the expression:
\begin{equation}
\label{eq:belUPMult}
    \rho_t^k(x) =\frac{\rho_{t-1}^k(x) \prod_{(a_t^k,y_t^k) \in (\rm a_t, \rm y_t )} P[y_t^k|a_t^k,x]}{\sum_{x'\in \mathcal{X}}\rho_{t-1}^k(x') \prod_{(a_t^k,y_t^k) \in (\rm a_t, \rm y_t )} P[y_t^k|a_t^k,x']}.
\end{equation} 
Similarly does Eve via the following belief update rule:
\begin{equation}
\label{eq:belUpEavMult}
    \rho_t^E(x)=\frac{\rho_{t-1}^E(x) \prod_{(a_t^k,z_t^k) \in (\rm a_t, \rm z_t)} Q[z_t^k|a^k_t,x]}{\sum_{x'\in \mathcal{X}}\rho_{t-1}^E(x') \prod_{(a_t^k,z_t^k) \in (\rm a_t, \rm z_t)} Q[z_t^k|a^k_t,x']},
\end{equation}
where ${\rm z}_t\triangleq(z_t^1,z_t^2,\ldots,z_t^K)$ denotes Eve's observations.

In this multi-agent case, we will further assume that each of the action sets $A^k$ also contains a ``no sensing action'' element, according to which the observations $y_t^k$ and $z_t^k$ are not generated. It is noted, however, that when a $k$-th legitimate agent selects this option, it can still update its belief using information from the other $K-1$ agents. We also consider that each agent can exit independently the sensing process, therefore, their communication graph may vary with time. To treat this general case, we consider as stopping time the time instance that the last agent exits the sensing process, i.e., $\tau=\max_k \tau_k$, with $\tau_k$ denoting the stopping time of each $k$-th agent. We also use notation $\gamma_{t,k}^L$ for the posterior error probability of each agent $k$ at time instant $t$. Similar to $\mathcal{OP}_1$, our goal is to find a collective agent policy ${\rm g}^C\triangleq({\rm g}_1,{\rm g}_2,\ldots,{\rm g}_k)$, where each ${\rm g}_k$ is obtained from \eqref{eq:policyDef}, that solves approximately the optimization problem:
\begin{align}
    \mathcal{OP}_2&:\nonumber\underset{{\rm g}^C}{\min} \,\, E[\tau]\\
    &\nonumber\quad\text{\text{s}.\text{t}.}\,\,\,\,\, \gamma_{\tau_k,k}^L \leq L\,\,\forall k, \,\,\gamma^E_t \geq E\,\,\forall t=1,2,\ldots,\tau.
\end{align}
%
\color{black}
Note that the error probability constraints can be relaxed using sample averaging as in~(\ref{eq:RelaxedConstraints}) for $\mathcal{OP}_1$.

\textit{Remark (The Importance of Inter-Agent Communication):} 
In our framework, we assume that each agent broadcasts its local  sensing information to neighboring agents aiming to support more informed inference. While this broadcasting incurs some communication overhead, it can significantly enhance the detection capabilities of all agents. Inter-agent information exchange is a well-established concept with numerous applications in modern intelligent wireless communication systems. For example, in spectrum sharing environments~\cite{marlSpectrumPoor}, agents share local observations with nearby peers to construct more accurate beliefs about the underlying channel states. In anomaly detection tasks, sensors collaborate by exchanging information to improve anomaly localization \cite{josephScalableAD}. Similarly, in edge caching systems, sharing learned representations of local observations and actions enables agents to refine popularity estimates, thereby increasing caching efficiency and network throughput \cite{edgeCachingMessages}. As it will demonstrated later on in Section~\ref{sec:results}, such communications can substantially reduce stopping times.
\color{black}
\section{Deep Neuroevolution Schemes for EAHT}\label{sec:NE}
In the section, we commence with the presentation of the application of NE to the considered centralized EAHT problem. Next, we present a novel NE-based method to deal with multi-agent POMDPs, which is then deployed to solve the considered decentralized EAHT problem. Finally, our NE-based method is extended to incorporate DNN pruning.

\subsection{Centralized EAHT}\label{sec:center_EAHT}
The policy DNN of an individual, which is needed in the NE formulation, is a mapping from beliefs to actions. We will use the Cooperative Synapse NE (CoSyNE) method~\cite{cosyne} to evolve a feed-forward policy DNN. The fitness function of a policy DNN $\theta$ is defined as follows:
\begin{equation}
\label{eq:Fitness}
    f(\theta)=\begin{cases}
    -A_E,\quad  A_E \geq 1-E \\

    \hat{E}_\tau^{-1},\quad \,\,\,  \text{otherwise}
    \end{cases},
\end{equation}
where $A_E\triangleq\hat{E} \max_t \max_{x} \pi_t^E(x)$ represents the average of the maximum Eve's belief value during an episode with $\hat{E}$ being the sample average, and $\hat{E}_\tau$ is the average horizon, both calculated from a large number $N_{\rm EP}$ of Monte Carlo episodes. It is noted that episodes in which Eve has large beliefs on some hypothesis are penalized. According to this definition, if a policy DNN cannot satisfy the privacy constraint, it is ``encouraged'' to do so; this is imposed from the first part of the fitness function $f(\theta)$. Otherwise, the DNN is ``encouraged'' to minimize the expected stopping time. Finally, individuals that satisfy the privacy constraint with the shortest stopping time are selected for mating. Similar to recent works on deep learning for AHT and related problems, e.g., \cite{ac-ad-noCost, stamatelisColab,stamatelis2023active,sacAD}, the policy DNN is only responsible for action selection. In this paper, we utilize a simple stopping rule, according to which termination takes place the first time $t$ for which holds $\gamma_t^L < L$. This stopping rule essentially handles the legitimate accuracy constraint, thus, it is unnecessary to include it in the fitness function calculation. A pseudocode describing our fitness function calculation is given in Algorithm~\ref{alg:fitnessCent}. 

\textit{Complexity Analysis:}
We will henceforth use the symbol $\boldsymbol{\theta}$ to denote the parameter vector concatenating all trainable weights $N_w$ of a policy DNN $\theta$. To this end, the CoSyNE algorithm maintains a \textit{population} of $L_{\rm pop}$ \textit{individuals} in a matrix $\boldsymbol{\Theta} \in \mathbb{R}^{L_{\rm pop} \times N_w}$, where each row corresponds to the weights of one individual with $N_w$ \textit{chromosomes}. Each $m$-th column of $\boldsymbol{\Theta}$ ($m=1,2,\ldots,N_w$) corresponds to each $m$-th \textit{subpopulation} of \textit{individuals}. This matrix is initialized randomly and the following steps are performed for each of the $N_{\rm gen}$ generations.
\begin{enumerate}
    \item \textit{Fitness evaluation:} Foremost, the fitness of all individuals comprising the population is evaluated, and then, those individuals are shorted according to it. For each individual $l$, each $l$-th row of $\boldsymbol{\Theta}$ ($l=1,2,\ldots,L_{\rm pop}$) is transformed  to a DNN and provided to Algorithm~\ref{alg:fitnessCent}.  Assuming that the forward pass time for a feed-forward DNN is $T_{\rm FP}$, then calculating the fitness for an individual requires $O(T_{\rm FP} T N_{\rm {EP}})$ of complexity.  Therefore, this step carries $O(L_{\rm pop} T_{\rm FP} T N_{\rm {EP}}+\log (L_{\rm pop}) )$  complexity.
    \item  \textit{Crossover and Mutation:} The top $\floor{L_{\rm pop}/4}$ rows of $\boldsymbol{\Theta}$ are used as parents to construct $\ceil{3L_{\rm pop}/4}$ offsprings, denoted by $\mathbf{\theta}^{\rm o}_{\ell}$ for $\ell=\ceil{3L_{\rm pop}/4},\ceil{3L_{\rm pop}/4}+1,\ldots, L_{\rm pop}$, through standard crossover and mutation mechanisms. Crossover combines the weights of two individuals and mutation adds Gaussian noise. The last $\ceil{3L_{\rm pop}/4}$ rows of $\boldsymbol{\Theta}$ are replaced by the offsprings. Mutation of the entire population requires $O(L_{\rm pop} N_w )$ of complexity, whereas crossover requires $O(L_{\rm pop}^2 N_w/16)$ of complexity.
    \item \textit{Permutation:} Each chromosome $[\boldsymbol{\Theta}]_{l,m}$ (i.e., each $(l,m)$-th element of $\boldsymbol{\Theta}$) is assigned the following permutation probability:
    \begin{equation*}
        p_{l,m}^{\rm perm}=1-\sqrt[N_w]{\frac{f_l}{f_{\rm max}}},
    \end{equation*}
    where $f_l$ is the individual's fitness function and $f_{\rm max}$ is the best fitness of the population. Then,  each chromosome is marked for permutation according to the above probability.
    For each $m$-th subpopulation $m=1,2,\ldots,N_{w}$, the marked chromosomes are shuffled. The complexity of this step is $O(N_w L_{\rm pop})$. 
    \end{enumerate}

Putting all above together, the computational complexity of the CoSyNE algorithm incorporated within the proposed single-agent EAHT is $N_{\rm gen} O((L_{\rm pop}(T_{\rm FP}  T N_{\rm EP}) + L^2_{\rm pop} N_w/16) )$. Note that, depending on implementation details (e.g., the type of data structures used), the exact complexity expression may differ.

\SetKwInput{KwData}{Input}
\setlength{\algomargin}{2em} 
\begin{algorithm}[!t]
\caption{Fitness for Centralized EAHT}\label{alg:fitnessCent}
\KwData{Individual DNN parameters $\theta$, thresholds $E$ and $L$, prior~$\pi_0(\mathcal{X})$, number of Monte Carlo episodes~$N_{\rm EP}$, and maximum horizon $T$.}
Set $A_E \gets 0$. \\
Set $\tau \gets 0$.\\
\For{$e_p=1,2,\ldots,N_{\rm EP}$}{
    Sample $x \sim \pi_0(\mathcal{X})$. \\
    Set $\pi_1^E(x)=\pi_1^L(x)=\pi_0(x)$ $\forall x\in\mathcal{X}$.\\
    \For{$t=1,2,\ldots,T$}{
        Choose action $a_t$ using the individual \\
        policy DNN $\theta$. \\
        Sample $y_t$ and $z_t$ from $P[y_t|a_t,x]$ and \\
        $Q[z_t|a_t,x]$, respectively. \\
        Update beliefs $\pi_t^L$ and $\pi_t^E$ using \\
        respectively (\ref{legBelUP}) and (\ref{eaveBelUP}). \\ 
        Set $\gamma_t^L \gets 1-\max_x \pi_t^L(x)$.\\
        \If{$\gamma_t^L \leq L $}{
            Break.
        }
    }
    Set $\tau \gets \tau+t$.\\
    Set $A_E \gets A_E + \max_t \max_x \pi_t^E(x)$.\\
}
Set $\tau \gets \tau/N_{\rm EP}$.\\
Set $A_E \gets $ $A_E/N_{\rm EP}$.\\
\If{$A_E \geq 1-E$}{
    \textbf{Output:} $-A_E$.
}
\textbf{Output:} $1/\tau$.
\end{algorithm}

\subsection{Decentralized EAHT}
We now present a novel dual-component deep NE approach for multi-agent cooperative tasks, which builds upon existing single-agent NE algorithms; to this end, we will use the previously mentioned CoSyNE algorithm~\cite{cosyne}, but other algorithms can be used as well. The proposed approach maintains all the previously highlighted NE benefits, and can be applied to tasks with multiple heterogeneous agents. Its first component is a feature extractor neural network that is utilized by all agents, and its second component consists of $K$ individual branches, one for each agent. The idea is to deploy the feature extractor weights to learn functions that will be used by all agents. The individual branches are then used to learn specific policies for each agent. Recall that the agents are in general heterogeneous, hence, they might have vastly different beliefs and action sets of different sizes. Moreover, some agents may have to remain inactive more often because their actions could cause very significant information leakage. For the latter reasons, the proposed approach uses individual branches. 

\begin{figure}[!t]
  \centering
  \hspace{10mm}
  \scalebox{0.8}{\includegraphics[width=0.5\textwidth]{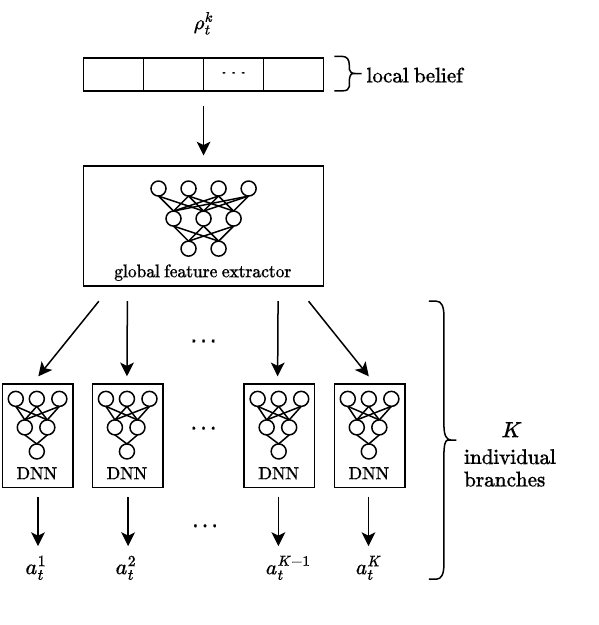}}
  \caption{The proposed neural network architecture for NE-based multi-agent cooperation. Each agent $k$ passes its belief to the global feature extractor $f$, and then, to its individual branch $b_k$ including a DNN for individual policy learning. The entire architecture can be evolved with typical NE algorithms, such as the deployed CoSyNE one here~\cite{cosyne}.}
  \label{fig:nn_decentralized}
\end{figure}

An individual with DNN parameters $\boldsymbol{\theta}$ can be split in the $K+1$ parts: $f$ and $b_1,b_2,\ldots,b_K$, where $f$ indicates the global extractor and each $b_k$ represents each $k$-th branch. 
The entire architecture is evolved as one network using CoSyNE~\cite{cosyne}. Evolutionary operations, such as crossover, are performed by one algorithm on the genes of the entire individual $\boldsymbol{\theta}$ and not on the separate branches, allowing us to maintain all the previously mentioned benefits of single-agent NE. \color{black}
During the deployment/testing phase, each agent is provided with the common feature extractor and its individual branch. The proposed neural network architecture for NE-based multi-agent cooperative tasks is illustrated in Fig.~\ref{fig:nn_decentralized}.

\label{subsec:DecEAHT}
In the proposed decentralized EAHT scheme, the fitness function of an individual is evaluated as follows. For each evaluation episode, the hypothesis is randomly sampled from $\pi_0(\cdot)$ and the beliefs of all agents are initialized. At each time instance $t$, each agent selects its action by passing the local belief through the feature extractor, and then, by forwarding the resulting output to its local branch. After $N_{\rm EP}$ Monte Carlo episodes take place, the fitness is computed according to~\eqref{eq:Fitness}, considering an appropriate adjustment to account for the decentralized stopping time, as defined in Section~\ref{sec:DedDef}. Each agent utilizes the stopping rule $\gamma_t^L < L$, as defined in the previous Section~\ref{sec:center_EAHT}. A pseudocode describing the fitness function calculation for the decentralized EAHT case is included in Algorithm~\ref{alg:fitnessDEc}.

\textit{Complexity Analysis:} By denoting with $T_{\rm FP}^G$ and $T_{\rm FP}^I$ the forward pass times for the global feature extractor neural network and the DNNs at the individual branches, respectively, the computational complexity of the fitness calculation for one individual is $O(K(T_{\rm FP}^G+T_{\rm FP}^I) T N_{\rm EP}))$. Since both the extractor and the individual branches are optimized as one joint structure, the time complexity required to construct a new generation of individual policy DNNs does not differ from that of the centralized method. Consequently, the total complexity of our decentralized NE-based optimization scheme is 
$N_{\rm gen} O(L_{\rm pop}(T_{\rm FP}^G+T_{\rm FP}^I)  T N_{\rm EP} + L^2_{\rm pop} N_w/16) )$.

\begin{algorithm}[!t]
\caption{Fitness for Decentralized EAHT}\label{alg:fitnessDEc}
\KwData{Individual DNN parameters $\theta$, thresholds $E$ and $L$, prior~$\pi_0(\mathcal{X})$, number of Monte Carlo episodes~$N_{\rm EP}$, maximum horizon $T$.}
Split $\theta$ to $f,b_1,b_2,\ldots,b_K$. \\
Set $A_E \gets 0$. \\
Set $\tau \gets 0$.\\
\For{$e_p=1,2,\ldots,N_{\rm EP}$}{
Sample $x \sim \pi_0(\mathcal{X})$. \\
Set $\rho_1^E(x)=\rho_1^k(x)=\pi_0(x)$ $\forall x\in\mathcal{X}$ and\\ $\forall k=1,2,\ldots,K$.\\
\For{$t=1,2,\ldots,T$}
{
\For{$k=1,2,\ldots,K$}{
Choose action $a_t^k$ using the  policy DNN\\ $b_k$ and the extractor $f$. \\ 
Sample $y^k_t$ and $z^k_t$ from $P[y_t^k|a^k_t,x]$ and\\ $Q[z_t^k|a^k_t,x]$, respectively.\\
}
\For{$k=1,2,\ldots,K$}{
Update beliefs $\rho_t^k$ using (\ref{eq:belUPMult}).\\
Set $\gamma_t^k \gets 1-\max_x \pi_t^k(x)$.\\
\If{$\gamma_t^k \leq L $}{Agent $k$ exits.}
}
Update $\rho^E_t$ using (\ref{eq:belUpEavMult}).\\
\If{All agents have exited}{Break.}

}

Set $\tau \gets \tau+t$.\\
Set $A_E \gets A_E + \max_t \max_x \rho_t^E(x)$.\\
}
Set $\tau \gets \tau/N_{\rm EP}$.\\
Set $A_E \gets A_E/N_{\rm EP}$.\\
\If{$A_E \geq 1-E$}{
\textbf{Output:} $-A_E$.
}
\textbf{Output:} $1/\tau$.
\end{algorithm}
\color{black}
\textit{Remark (The Role of the Global Extractor):} While we use the term ``global'' to refer to the feature extractor $f$, this operator actually processes only the local beliefs $\rho_t^k$ for each agent $k$. To this end, copies of the same parameters of $f$ are shared by all agents. This is the intention behind the term ``global,'' which is used to learn common operations that will be used by all agents to improve efficiency. Parameter sharing is generally very successful in multi-agent DRL \cite{multiACMixed,scalingUpMARL} and is preferred over fully independent DNNs. This motivated us to adopt it herein in our NE framework. 
\color{black}
\subsubsection{Joint NE and Pruning}
In this section, we present a decentralized EAHT scheme that builds upon Algorithm~\ref{alg:fitnessDEc} combining NE and pruning~\cite{stateNetPrun,marlPrune}. In particular, the proposed scheme comprises two distinct steps: \textit{i}) a joint optimization and pruning step; and \textit{ii}) a fine-tuning step for the pruned solution, as shown in Algorithm~\ref{alg:joint_evolution_pruning}. Each step employs a separate run of the CoSyNE algorithm to achieve its objectives.

In the first step, we initialize a population of dense DNNs, where each network comprises the global feature extractor $f$ and the $K$ individual branches $b_1,b_2,\ldots,b_K$, as described previously. During each fitness function evaluation, every layer of the candidate DNN undergoes unstructured weight-level pruning by a predefined pruning percentage $p_i$. To this end, redundant weights are set to zero, and the pruned network is subsequently evaluated following the procedure as the previous unpruned decentralized NE-based scheme. After the evaluation round, the top-performing individuals are selected and combined (mated) to produce the next generation of candidate solutions. This evolutionary process is repeated over multiple generations, and as pruning is applied iteratively, the networks are typically pruned beyond the initial pruning percentage $p_i$. The output of this step, denoted as $\theta^*$, is a sparsely structured network optimized for both performance and efficiency. This extensive pruning effect will be experimentally verified.

In the second step, the sparse structure of $\theta^*$ is preserved, and the focus shifts to fine-tuning its nonzero weights. A population of networks is initialized, each retaining the structure of $\theta^*$, with unnecessary parameters masked to zero. For each individual, the nonzero weights of $\theta^*$ are copied and perturbed with small-magnitude Gaussian noise to introduce diversity. 
The CoSyNE algorithm is then applied to this newly constructed population, enabling standard evolutionary refinement of the pruned solution. Individual evaluations in this step follow the same procedure outlined in the previous unpruned scheme.  

The overall procedure for this decentralized EAHT scheme implementing joint NE and pruning, which is summarized in Algorithm~\ref{alg:joint_evolution_pruning}, leverages Algorithm~\ref{alg:fitnessDEc} to evaluate candidates. It noted that our proposed joint NE-based optimization and pruning framework is general and can be applied to any genetic algorithm of choice besides the CoSyNE algorithm we are using in this paper.

\textit{Complexity Analysis:} Pruning operations for a DNN with $N_w$ learnable weights can be achieved with linear time complexity. Therefore, for the first step of our decentralized EAHT scheme with joint NE and pruning, the complexity expression becomes $N_{\rm gen} O(L_{\rm pop}(T_{\rm FP}^G+T_{\rm FP}^I )N_w  T N_{\rm EP} + L^2_{\rm pop} N_w/16) )$. For the second step, the complexity expression is  $N_{\rm gen} O(L_{\rm pop}(T_{\rm FP}^{G'}+T_{\rm FP}^{I'})  T N_{\rm EP} + L^2_{\rm pop} N_w'/16) )$,  where $N_w'$ represents the number of non-zero weights of $\theta^*$, whereas $T_{\rm FP}^{G'}$ and  $T_{\rm FP}^{G'}$ denote the forward pass time of the pruned extractor and that of the individual branch DNNs, respectively.
Since a significant number of weights will be pruned, we can safely assume that $N_w' \ll N_w$, $T_{\rm FP}^{G'} < T_{\rm FP}^{G}$, and $T_{\rm FP}^{l'}< T_{\rm FP}^{l}$, implying that the $N_{\rm gen} O(L_{\rm pop}(T_{\rm FP}^{G'}+T_{\rm FP}^{I'})  T N_{\rm EP} + L^2_{\rm pop} N_w'/16) )$ term does not need to be included in the complexity expression.

\begin{algorithm}[!t]
\caption{Joint NE and Pruning}\label{alg:joint_evolution_pruning}
\KwData{Pruning percentage $p_i$, number of generations~$N_{\rm gen}$, population size $L_{\rm pop}$, and~noise variance $\sigma^2$.}

\textbf{Step 1: Joint Optimization and Pruning}\\
Initialize a population of $L_{\rm pop}$ dense DNNs.\\
\For{generation $g = 1, 2, \ldots, N_{\rm gen}$}{
    \For{each individual in the population}{
        Apply unstructured pruning with percentage\\ $p_i$ to each layer.\\
        Evaluate the pruned network using\\ Algorithm~\ref{alg:fitnessDEc}.\\
    }
    Select top-performing individuals for mating.\\
    Generate offsprings through genetic operations.\\
}
Let $\theta^*$ be the DNN of the best pruned individual\\ from the final generation.\\

\textbf{Step 2: Fine-Tuning the Pruned Solution}\\
Initialize a new population of $L_{\rm pop}$ DNNs\\ with the sparse structure of $\theta^*$.\\
\For{each individual in the population}{
    Copy nonzero weights from $\theta^*$.\\
    Add the Gaussian noise $\mathcal{N}(0, \sigma^2)$ to the copied\\ weights.\\
}
\For{generation $g = 1, 2, \ldots, N_{\rm gen}$}{
    \For{each individual in the population}{
        Evaluate the individual using Algorithm \ref{alg:fitnessDEc}.\\
    }
    Select top-performing individuals for mating.\\
    Generate offsprings through genetic operations.\\
}
\textbf{Output:} Final fine-tuned sparse DNN $\theta^*$.
\end{algorithm}

\section{Numerical Results and Discussion}\label{sec:results}
In this section, we present performance evaluation results for our NE-based single- and multi-agent EAHT schemes, considering an anomaly detection scenario over wireless sensor networks. Different values for the thresholds $L$ and $E$ as well as for the number of sensors, $S$, have been considered. Two different statistical sensor models commonly adopted in recent literature have been implemented~\cite{DRL-AHT-AD-journal,decHTKohen}. \textcolor{black}{In addition, we considered an anomaly detection application where sensors transmit their data over Ricean fading channels, as well as a radar-assisted target detection application similar to~\cite{chernoff-radar}.}

\begin{table}[!t]
    \centering
        \caption{Flipping probabilities for each sensor's three distinct access actions.}
    \begin{tabular}{|c|c|c|}
    \hline
        Sensor Access Action Number& $P_{\rm flip}^L$ & $P_{\rm flip}^E$ \\
        \hline\hline
         1& 0.125 & 0.125\\
         \hline
         2 & 0.2 &0.4 \\
         \hline
         3 &0.25 &0.45 \\
         \hline
    \end{tabular}
    \label{tab:flipProbs}
\end{table}

\subsection{NE Implementation}

The proposed single-agent NE scheme uses a feed-forward DNN with $2$ hidden layers each comprising $n_{\rm h}=200$ weights, whereas the proposed multi-agent algorithm utilizes a feature extractor with $2$ hidden layers of $n_{\rm f}=300$ hidden weights and $K$ branches (each corresponding to one of the $K$ agents/sensors), each including a $2$-layer DNN with $n_{\rm b}=300$ weights per layer. The mutation probability, $p_{\rm mut}$, was set to the relatively high value $0.5$ to ensure sufficient exploration, whereas the standard deviation of the mutation, $\sigma_{\rm mut}$, was given the value to $0.6$. The population size was set to $L_{\rm pop}=50$ and we evolved the DNNs over $N_{\rm gen}=50$ generations. For each individual, the fitness function was evaluated over $N_{\rm EP}=100$ episodes, and, for the decentralized EAHT scheme with pruning, we have set $p_i=0.2$. The obtained results of all learning agents were further averaged by running the respective algorithms for $20$ different initialization seeds. All DNNs were trained on a GeForce RTX $3080$ GPU with $32$~GB memory.
\subsection{Benchmark Schemes}
For the single-agent problem, we have implemented two benchmark AHT strategies that ignore the existence of Eve: the Chernoff test~\cite{ChernoffHT,controlledSensForMH} and a myopic Extrinsic Jensen-Shannon (EJS) divergence maximization strategy~\cite{nagshvarEJS}. In addition, since the majority of recent work on learning-based AHT uses deep actor critic algorithms, e.g., \cite{josephScalableAD,decHTKohen,stamatelisColab,sacAD,stamatelis2023active}, we have considered one such algorithm with appropriately modified reward structures. We have particularly simulated the performance of two PPO DRL algorithms rewarded for error minimization, such as the one in~\cite{stamatelisColab,stamatelis2023active}, and for confidence maximization, similar to the one in~\cite{KartrikDQN}. For these algorithms, if the privacy constraint failed, a large penalty was reached and, consequently, the episode was terminated. Similar penalty-based rewards have been used in related POMDPs, e.g., in~\cite{activePUT}.
For this DRL approach, larger DNNs than for the proposed NE-based schemes were used, in particular, an actor and a critic with $2$ hidden layers each consisting of $300$ learnable weights. \textcolor{black}{Besides PPO, we also used an Advantage Actor Critic (A2C)~\cite{asynchronousDRL} algorithm and a DQN \cite{mnih2015humanlevel} with the first reward structure.}

\color{black}
Decentralized POMDPs are known to require at least exponential complexity~\cite{decPOMDPComplexity}, hence, it is extremely difficult to use mathematical methods for the considered multi-agent problem. For this reason, we have focused on baseline learning-based algorithms and used two DRL algorithms with individual actors and a global larger critic similar to~\cite{decHTKohen,stamatelisColab,multiACMixed}. More specifically, an  Actor Critic (AC) algorithm and a PPO extension of the MADDPG structure \cite{multiACMixed} have been developed\footnote{\textcolor{black}{Besides EAHT, similar algorithms have ben deployed in a wide variety of wireless applications, including caching \cite{MADDPGCaching}, dynamic spectrum access~\cite{stamatelisHarmonicAnnealedPruning}, and power allocation \cite{MADDPGPowerControl}, making them representative powerful benchmarks.}}. In the implementation of these benchmarks, when the privacy constraint was satisfied, they were rewarded for error minimization, otherwise, a large penalty was received. Apart from the penalty, the training was nearly identical to state-of-the-art DRL approaches for multi-agent active sensing~\cite{decHTKohen,stamatelisColab}. We also used two counterparts with gradual unstructured pruning, where the sparsity levels gradually increased according to a polynomial schedule \cite{HanPrun}. More advanced pruning methods, like the recent one in~\cite{stamatelisHarmonicAnnealedPruning}, are left for future work.


\color{black}
\subsection{Results for Centralized EAHT}\label{subsec:centralizedResults}
A number of $S$ independent and identical sensors were tasked to detect anomalies in their proximity~\cite{DRL-AHT-AD-journal,josephScalableAD}. We have assumed that any number of sensors can be near an anomaly, hence, there were in total $2^S$ possible hypotheses. At each time instance $t$, the single agent probed one sensor and received the following binary observation:
\begin{equation}
    \label{yDef}
    y_t=\begin{cases}
        s, \quad\quad\,\,\,\, \text{with probability } 1-P_{\rm flip}^L \\ 
        1-s,  \quad \text{with probability} P_{\rm flip}^L \\
    \end{cases},
\end{equation}
where $s$ is a binary number corresponding to the sensor's state (whether it is near an anomaly or not) and $P_{\rm flip}^L$ is the flipping probability. A similar expression held for Eve's observation $z_t$, whose flipping probability is denoted by $P_{\rm flip}^E$. Note that binomial sensor models have been assumed in various relevant references, e.g., \cite{DRL-AHT-AD-journal,josephScalableAD,stamatelis2023active,ac-ad-noCost}. We have further assumed that the single agent can access each sensor with three different actions, each corresponding to one of the three different flipping probability values. Therefore, the total actions available to the agent were $3S$. The three different flipping probability values and the respective three distinct sensor access actions are listed in Table~\ref{tab:flipProbs}. 

\begin{figure*}
    \centering
    \begin{subfigure}{.33\textwidth}
        \centering
        \includegraphics[width=\textwidth]{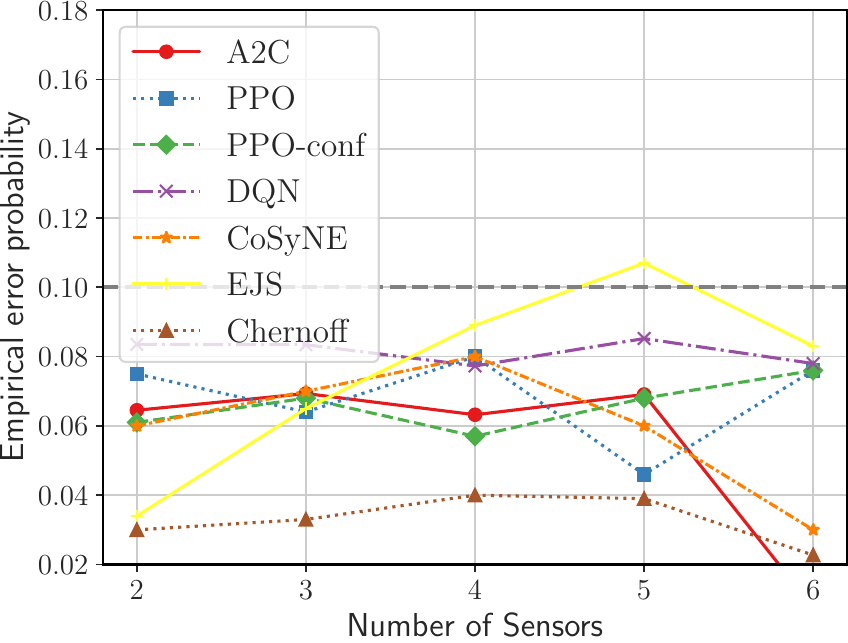}
        \caption{Legitimate error probability: $E=0.3$.}
    \end{subfigure}\hfill
    \begin{subfigure}{.33\textwidth}
        \centering
        \includegraphics[width=\textwidth]{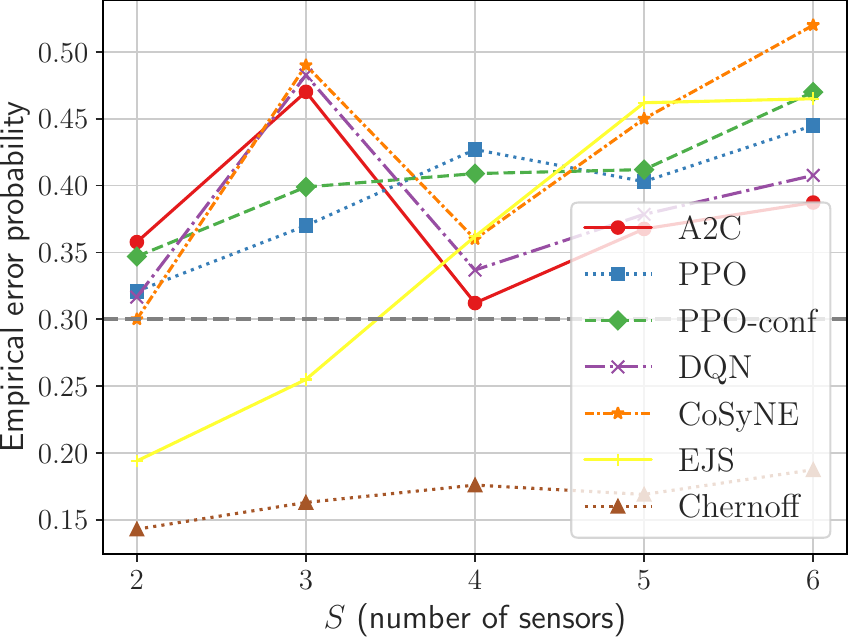}
        \caption{Eve's error probability: $E=0.3$.}
    \end{subfigure}\hfill
    \begin{subfigure}{.33\textwidth}
        \centering
        \includegraphics[width=\textwidth]{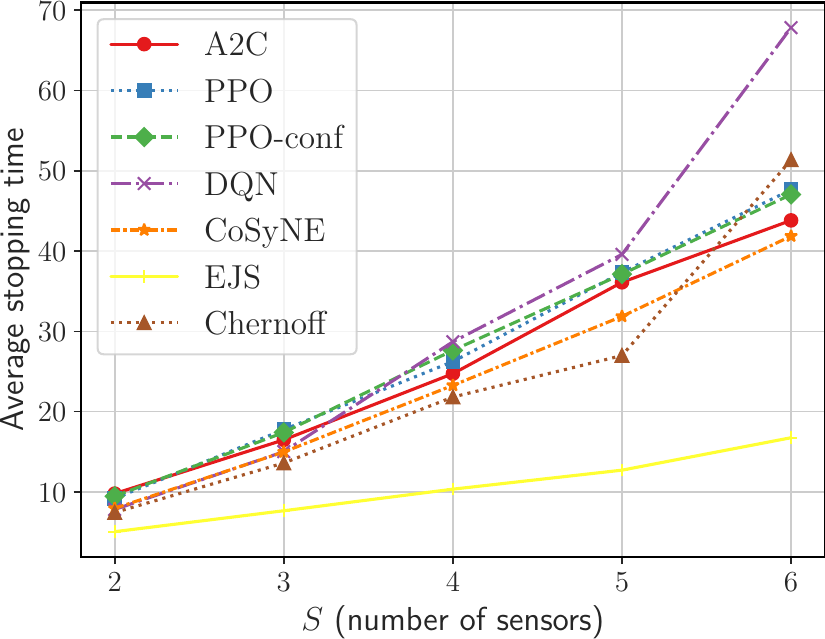}
        \caption{Average episode stopping time: $E=0.3$.}
    \end{subfigure}\\
    \begin{subfigure}{.33\textwidth}
        \centering
        \includegraphics[width=\textwidth]{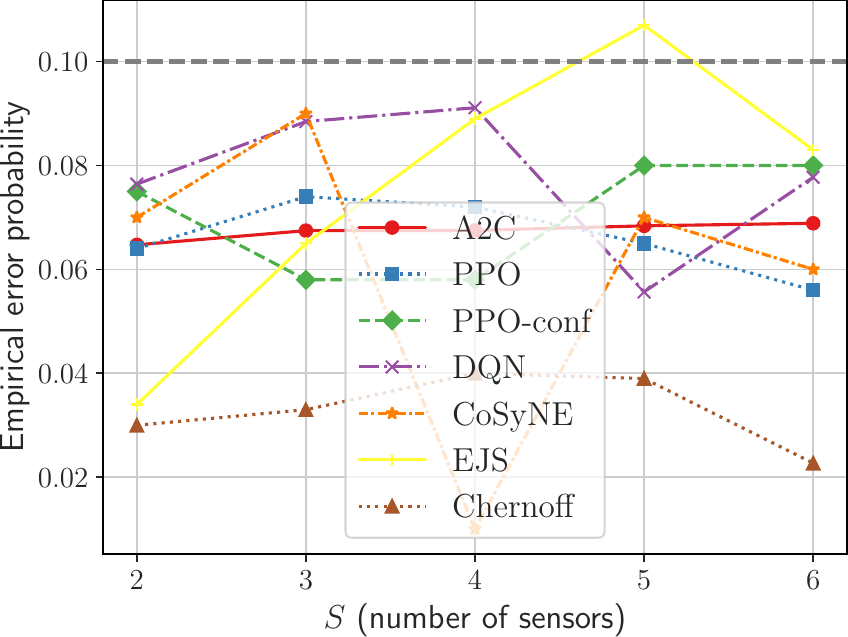}
        \caption{Legitimate error probability: $E=0.4$.}
    \end{subfigure}\hfill
    \begin{subfigure}{.33\textwidth}
        \centering
        \includegraphics[width=\textwidth]{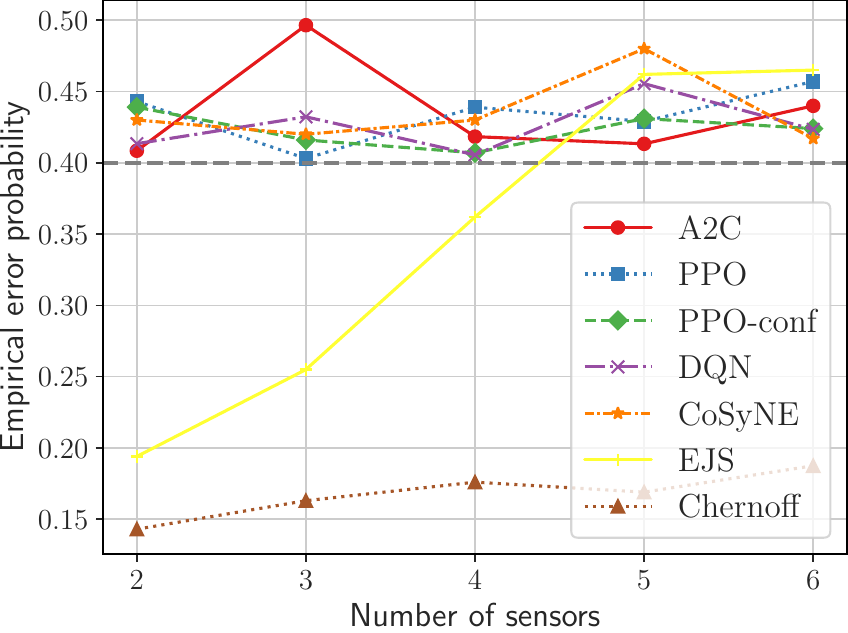}
        \caption{Eve's error probability: $E=0.4$.}
    \end{subfigure}\hfill
    \begin{subfigure}{.33\textwidth}
        \centering
        \includegraphics[width=\textwidth]{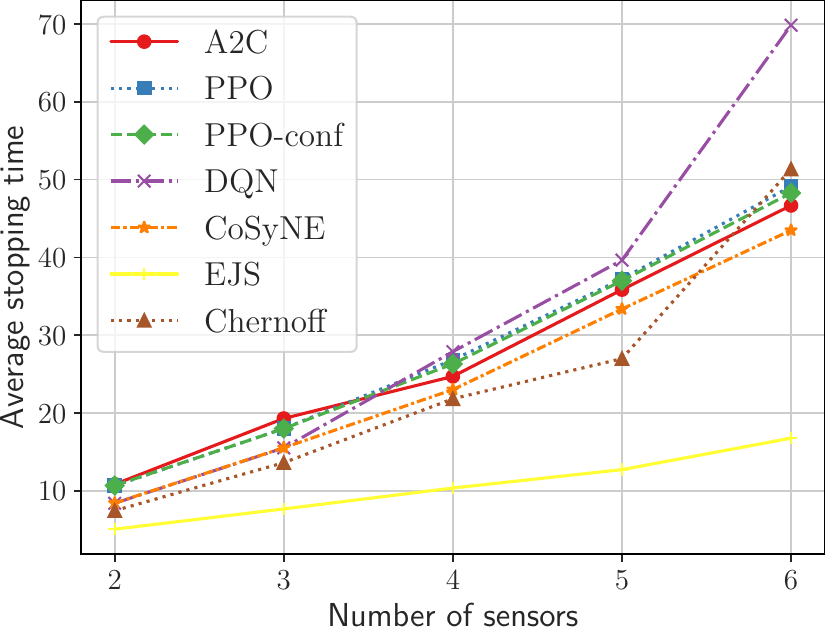}
        \caption{Average episode stopping time: $E=0.4$.}
    \end{subfigure}
    \caption{Error probability performance for the centralized (single-agent) EAHT problem considering \eqref{yDef}'s binomial sensor model.}\label{fig:single-binomial}
\end{figure*}
In the performance results illustrated in Fig. ~\ref{fig:single-binomial}, we have set $L=0.1$ and considered two values for $E$, namely $E=0.4$ and $0.3$. In addition, the number of available sensors $S$ was varied between $2$ to $6$. It can be first observed that the legitimate error probability is lower than the threshold value $L$ for all approaches besides the EJS benchmark. Interestingly, the proposed NE-based EAHT approach and the PPO benchmarks lead to substantially higher error probability on Eve's side. On the other hand, the conventional approaches cannot satisfy the privacy constraints, resulting in a large margin from the latter best schemes. It can also be seen that, for $S<4$, Eve's error probability when running these benchmarks is always less than $0.3$. In addition, for some experiments, Eve's error probability was more than $50\%$ smaller than the desired threshold. It can be also seen that, as expected, there is a trade-off between the episode stopping time and the privacy objective, since the proposed NE-based and PPO methods need to perform a few more sensing actions. It is finally shown, that the CoSyNE algorithm achieves a shorter average stopping time than the PPO benchmarks in all simulated investigations. In fact, in some simulations' settings, the stopping time of the CoSyNE algorithm was $20\%$ shorter than the stopping time with the PPO algorithms. Moreover, for $S=6$, CoSyNE terminates faster than the Chernoff test, which by design ignores the existence of the eavesdropper. 

\begin{figure}
    \centering
    \includegraphics[width=0.65\linewidth]{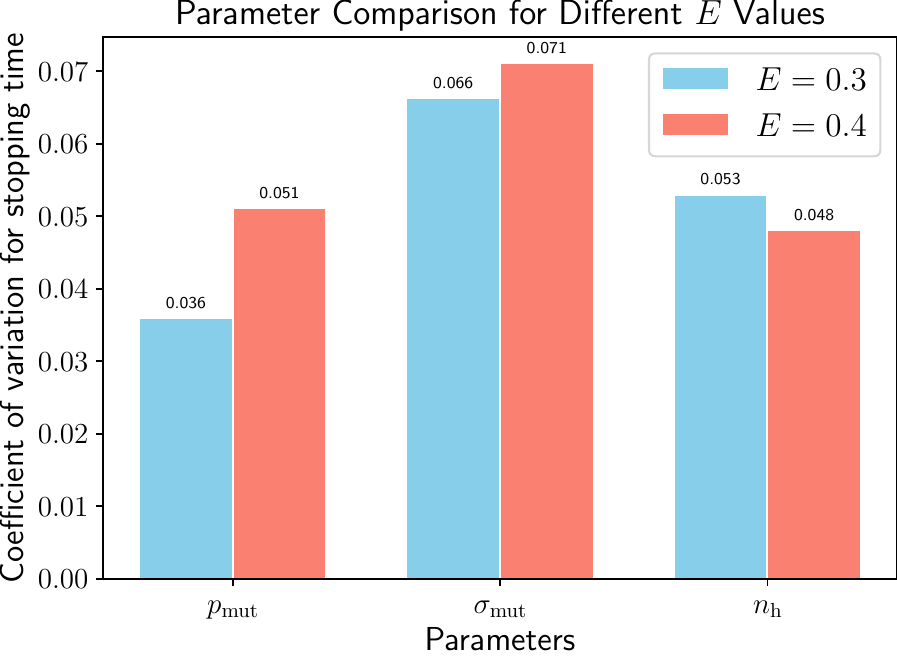}
    \caption{\textcolor{black}{Coefficients of Variation (CVs) of the stopping time objective for the single-agent NE-based EAHT scheme, considering $S=6$ sensors and variations of hyperparameters.}}
    \label{fig:SensitivitySingleAgent}
\end{figure}

\color{black}
To investigate the robustness of the CoSyNE optimizer, we conducted a sensitivity study under deviations in the hyperparameters focusing on the larger environment with the $S=6$ sensors. For each hyperparameter, we varied its value, keeping the rest fixed. The training and testing procedures were repeated for each configuration in order to examine the effect that each hyperparameter has on the algorithm's performance. We varied the mutation probability $p_{\rm mut}$ from $0.3$ to $0.6$, the standard deviation $\sigma_{\rm mut}$ from $0.4$ to $0.8$, and the hidden size of the DNN $n_{\rm h}$ from $150$ to $300$. The generated results demonstrated that, for each case, the CoSyNE optimizer can discover policies that satisfy the constraints, while reaching ``good'' stopping times. The Coefficients of Variation (CVs) for the stopping time are depicted in Fig.~\ref{fig:SensitivitySingleAgent}, clearly demonstrating the stability of our scheme. It is particularly shown that the CVs are consistently significantly smaller than $0.1$, signifying very good robustness; the error probabilities are omitted from the presentation due to space constraints.

\begin{figure}
    \centering
    \includegraphics[width=0.65\linewidth]{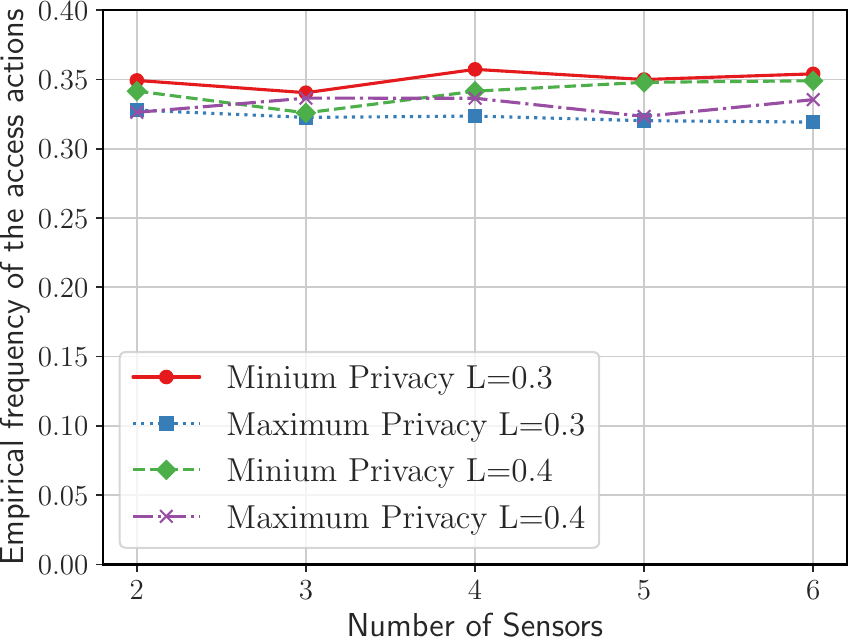}
    \caption{\textcolor{black}{The frequency of the first and third sensor access actions of the final evolved policies considering \eqref{yDef}'s binomial sensor model.}}
    \label{fig:vizPolicies}
\end{figure}
In Fig. \ref{fig:vizPolicies}, we plot the probabilities with which the optimized policies select the access action with the minimum privacy (both flipping probabilities are set to $0.125$) and the maximum privacy ($P^L_{\rm flip}=0.25, P_{\rm flip}^E=0.45$) access modes. It is apparent that the solution is not trivial (e.g., select only maximum privacy, or quickly detect the hypothesis ignoring the $E$ values). In fact, a balance of all three access/protection levels is required to ensure secure and reliable inference. Interestingly, leakage to the eavesdropper is sometimes accepted in order to form initial estimates, and when quality beliefs are formed by agent $L$, less informative actions can be selected. It can be also seen that, for larger $E$ values, the third action that maximizes privacy is taken a little more often.

\color{black}
To further validate the effectiveness of our approaches, we have simulated a second sensor model including Gaussian observations, similar to \cite{decHTKohen,sacAD}. According to this model, the observations returned by a probed sensor are given by:
\begin{equation}
    \label{eq:GaussOBS}
    y_t \sim \begin{cases}
       \mathcal{N}\left(1,\sigma_l^2\right), \quad \text{if the sensor is near an anomaly.}\\
        \mathcal{N}\left(0,\sigma_l^2\right), \quad \text{otherwise}.
       
    \end{cases}
\end{equation}
Again, a similar expression held for the Eve's observations $z_t$ with variance $\sigma_e^2$. Like in the previous binomial model, we assumed three sensing modes for each sensor. The considered values of $\sigma_l^2$ and $\sigma_e^2$ for each mode are included in Table~\ref{tab:variances}.

\begin{figure*}
    \centering
    \begin{subfigure}{.33\textwidth}
        \centering
        \includegraphics[width=\textwidth]{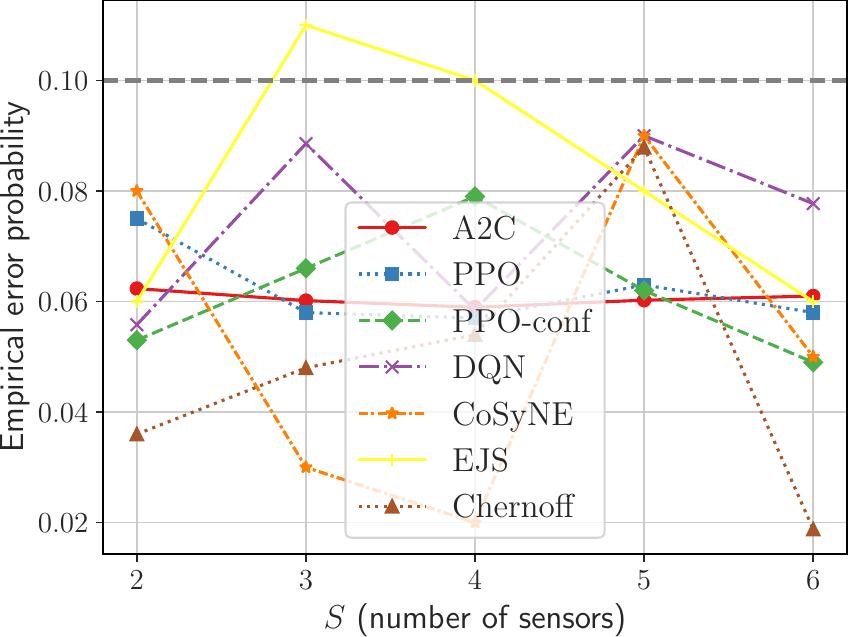}
        \caption{Legitimate error probability: $E=0.3$.}
    \end{subfigure}\hfill
    \begin{subfigure}{.33\textwidth}
        \centering
        \includegraphics[width=\textwidth]{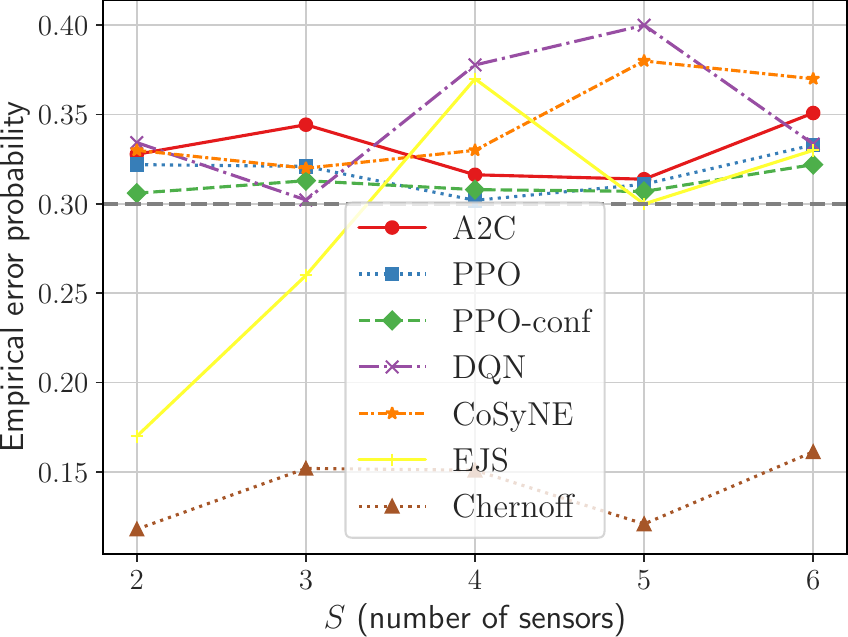}
        \caption{Eve's error probability: $E=0.3$.}
    \end{subfigure}\hfill
       \begin{subfigure}{.33\textwidth}
        \centering
        \includegraphics[width=\textwidth]{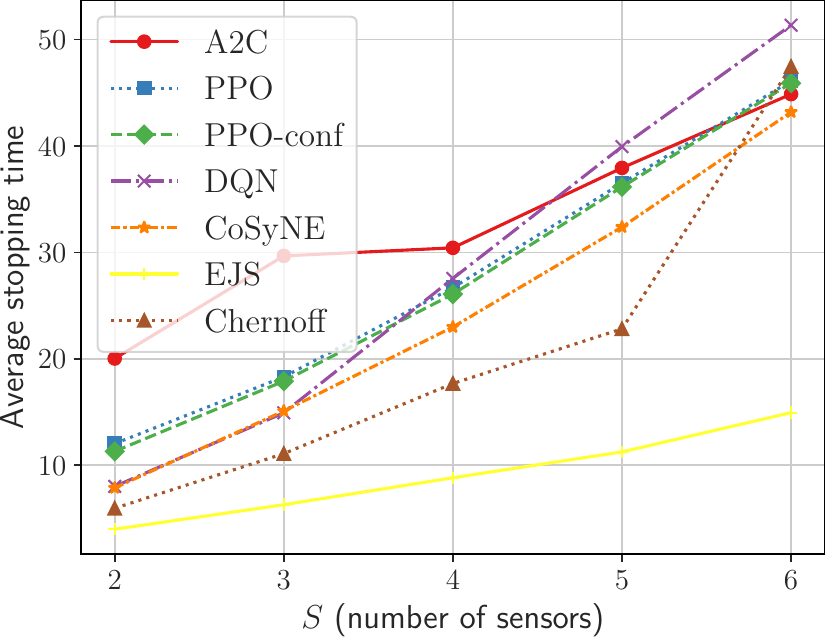}
        \caption{Average episode stopping time: $E=0.3$.}
    \end{subfigure}\hspace{1.25cm}
    \caption{Similar to Fig.~\ref{fig:single-binomial}, but for the Gaussian sensor model in \eqref{eq:GaussOBS}.}\label{fig:single-gaussian}
\end{figure*}

By repeating the experiments of Fig.~\ref{fig:single-binomial} for the Gaussian sensor model and $E=0.3$, it can be observed from the obtained results within Fig.~\ref{fig:single-gaussian} that again CoSyNE outperforms all benchmarks. Evidently, the EJS and the Chernoff algorithms cannot consistently satisfy the privacy constraint, and EJS misses the accuracy constraints. On the other hand, both the CoSyNE and the PPO algorithms satisfy the constraints, with the former achieving noticeably shorter stopping times.

All in all, it can be concluded from the results, for both sensor models in Figs.~\ref{fig:single-binomial} and~\ref{fig:single-gaussian}, that the classic benchmarks cannot satisfy the privacy and accuracy constraints in most settings. Interestingly, our proposed NE framework does meet those constraints in all settings, while achieving shorter stopping time than the considered modified DRL benchmarks with appropriate penalized reward signals.
\begin{table}[!t]
    \centering
    \begin{tabular}{|c|c|c|}
    \hline
         Sensor Access Action Number& $\sigma_l^2$ & $\sigma_e^2$  \\
         \hline
         1& 0.25 &0.25 \\
         \hline
         2&0.5& 1.25\\
         \hline
         3&1&2.5\\
         \hline
    \end{tabular}
    \caption{Variances for each sensor's three distinct access actions.}
    \label{tab:variances}
\end{table}

\color{black}
\subsubsection{Robustness Against DNN-Based Eavesdroppers}\label{subsec:learningeave}
We now examine how our evolved policies can deal with eavesdroppers having learning capabilities. Focusing on the binomial observations of~(\ref{yDef}) with $E=0.3$, we have collected a large training dataset of $80000$ episodes using our final policies, and another test set of $10000$ episodes. Each data point contains a sequence of actions $a_t$, observations $z_t$, and a label for the true hypothesis. We have trained different machine learning classifiers to verify that our method can trick a diverse set of adversaries. Due to the fact that trajectories have different lengths, we have considered two Recurrent Neural Networks (RNNs) with $3$ hidden layers of $300$ units, followed by a final softmax-activated output layer for classification. The RNNs were particularly a Long Short Term Memory Network (LSTM)~\cite{LSTM} and a Gated Recurrent Unit (GRU)~\cite{GRUPaper}, with both being bidirectional enabling greater representation capabilities. These networks were developed using the pytorch framework~\cite{pytorch} and optimized with the Adam optimizer~\cite{adam} using a learning rate of $2.5 \times 10^{-4}$. We have also used a decision tree classifier which pads all sequences to the same length, i.e., max-padding.

\begin{figure}
    \centering
\includegraphics[width=0.65\linewidth]{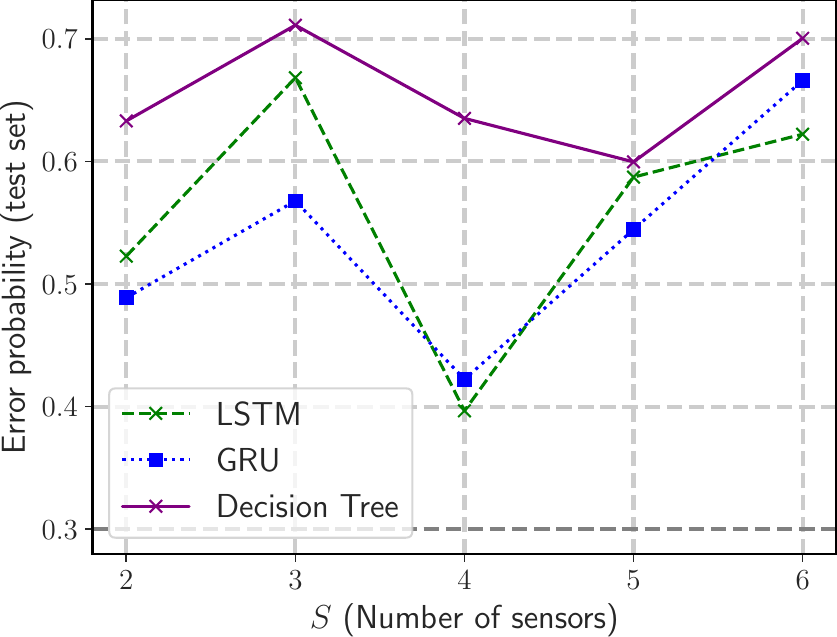}
    \caption{\textcolor{black}{Test set error probability estimates of various DNN-based eavesdroppers.}}
    \label{fig:LearningEve}
\end{figure}

It is important to note that an eavesdropper equipped with a large, high-fidelity dataset that closely matches the actual system represents a particularly powerful adversary; this effectively models a worst-case scenario. In many real-world deployments, such strong adversaries may not exist due to limited data access or mismatched system knowledge. Nonetheless, our proposed policies demonstrate strong resilience even under this pessimistic assumption. As illustrated in Fig.~\ref{fig:LearningEve}, they are capable of misleading DNN-based eavesdroppers effectively across a wide range of settings.

\subsubsection{Incorrect Knowledge of Observation Models}\label{subsec:missmatch}
So far, we have considered that both $P[\cdot]$ and $Q[\cdot]$ are correctly estimated from a large dataset, as well as that the deployment conditions of our policies are identical to training. We will henceforth examine how our approach and the DRL benchmarks perform when $P[\cdot]$ and $Q[\cdot]$ are only crude approximations of the true kernels $P_{\rm true}[\cdot]$ $ Q_{\rm true}[\cdot]$, respectively. We have particularly focused on the larger binomial sensor environment, testing the trained policies of Fig.~\ref{fig:single-binomial} for $S=6$ sensors. We assumed that the agent updates its belief using the flipping probabilities of Table~\ref{tab:flipProbs}, however, the probabilities of the testing environment have been slightly perturbed being non-constant. More specifically, the legit flipping probability at each time instance $t$ was sampled from the uniform distribution in $[0.85P_{\rm flip}^L,1.15P_{\rm flip}^L]$. It was also assumed that the agent underestimates the capabilities of the eavesdropper, whose flipping probability at each time instant $t$ was set to lie uniformly in $[0.7 P_{\rm flip}^E,0.9 P_{\rm flip}^E]$.

As shown in Table \ref{tab:gener}, our proposed NE-based approach, along with the PPO and DQN agents, successfully satisfies both the utility and privacy constraints. However, it can be seen that the A2C agent exhibits a slight violation of the privacy constraint. Notably, our method continues to achieve the shortest stopping times while maintaining robust performance under model mismatch. This highlights its strong generalization capability making it particularly well-suited for deployment in dynamic or uncertain environments.

\begin{table}[]
    \centering
    \begin{tabular}{|c|c|c|c|}
    \hline
        Algorithm& Legit. Error Prob. & Eav. Error Prob.  & Aver. Stop. Time \\
         \hline
         \hline
         A2C& 0.0965 & \textbf{0.2776} & 46.633 \\ 
         PPO-conf& 0.0875  & 0.3781 & 48.15 \\
         PPO & 0.0544 & 0.3656 & 48.101 \\ 
         DQN & 0.0733 & 0.314 & 73.87 \\
         \hline
         CoSyNE& 0.0621 & 0.3415 & 42.056\\ 
         \hline
    \end{tabular}
    \caption{\textcolor{black}{Results for the case of incorrect knowledge of flipping probabilities and $S=6$ sensors.}}
    \label{tab:gener}
\end{table}

\subsubsection{Further Wireless System Applications}\label{subsec:application}
To further demonstrate the effectiveness of our approach, we have explored two additional realistic applications: an extension of the current sensor network scenario incorporating multipath fading conditions, and radar-based object detection.

\paragraph{Sensor Networks under Ricean Fading Channels}  
We have considered a system with \( S=3\) sensors, each capable of detecting anomalies and broadcasting symbols \(x_{t,s}\) according to the following rule:
\begin{equation}
x_{t,s} = 
\begin{cases}
    1, & \text{if } s = 1 \\
    -1, & \text{otherwise}
\end{cases}.
\end{equation}
At each time step \( t \), the agent \( L \) selects both a sensor and a transmit power level \( P_t \in \mathcal{P} \), where \( \mathcal{P} \) is a discrete set of allowable power levels. The baseband received signal at agent \( L \) is mathematically modeled as follows:
\begin{equation}
\hat{x}_{t,s}^L = \sqrt{P_t} h^L_{t}x_{t,s} + n,
\end{equation}
where \( h_{t}^L \) represents the complex Ricean fading channel gain coefficient between the selected sensor and agent \( L \), and \( n \) is the Additive White Gaussian Noise (AWGN). The agent is then assumed to apply a hard decision rule according to which it decodes the received signal as \( y_t = 1 \) if \( \operatorname{Re}(\hat{x}_{t,s}^L) > 0 \), and \( y_t = 0 \) otherwise. The flipping probabilities for each sensor and the power level have been empirically estimated using extensive Monte Carlo simulations.

A similar observation model has been considered for the eavesdropper, who receives the signal through an independent fading channel \( h_{t}^E \), which has been assumed to be characterized by a weaker Line-of-Sight (LoS) component. We have specifically used \( \kappa^L = 5\,\mathrm{dB} \) as the Ricean factor for the legitimate agent and \( \kappa^E = -2\,\mathrm{dB} \) as that for the eavesdropper. 
The set of  the set of available power levels was \( \mathcal{P} = \{-20\,\mathrm{dB}, -10\,\mathrm{dB}, 0\,\mathrm{dB}\} \).

\begin{figure*}
    \centering
    \begin{subfigure}{.33\textwidth}
        \centering
        \includegraphics[width=\textwidth]{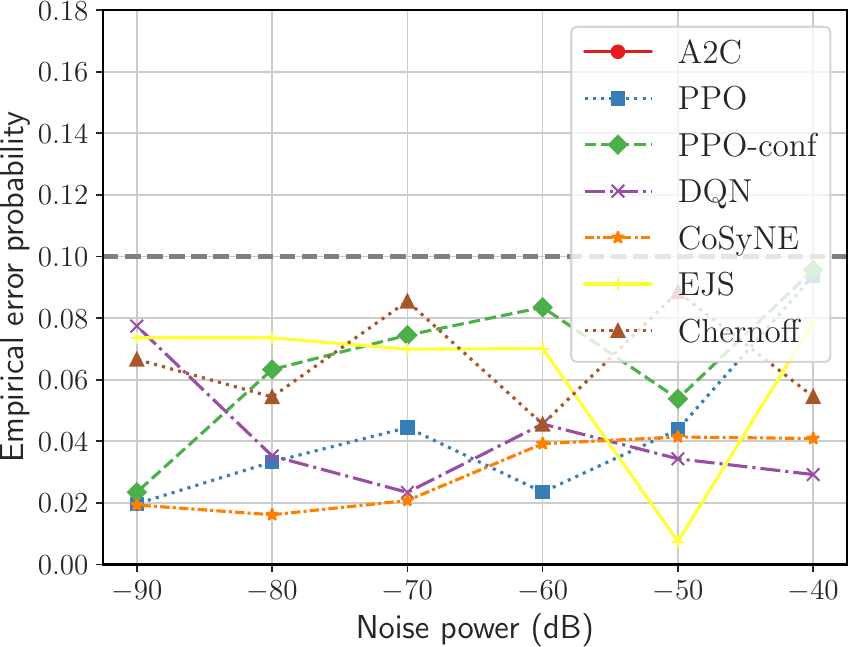}
        \caption{Legitimate error probability: $E=0.3$.}
    \end{subfigure}\hfill
    \begin{subfigure}{.33\textwidth}
        \centering
        \includegraphics[width=\textwidth]{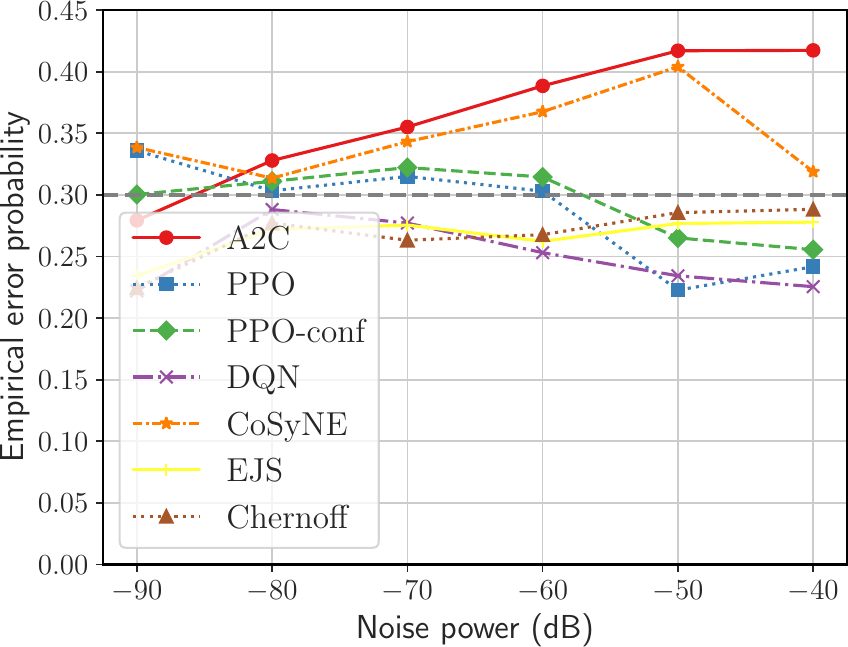}
        \caption{Eve's error probability: $E=0.3$.}
    \end{subfigure}\hfill
       \begin{subfigure}{.33\textwidth}
        \centering
        \includegraphics[width=\textwidth]{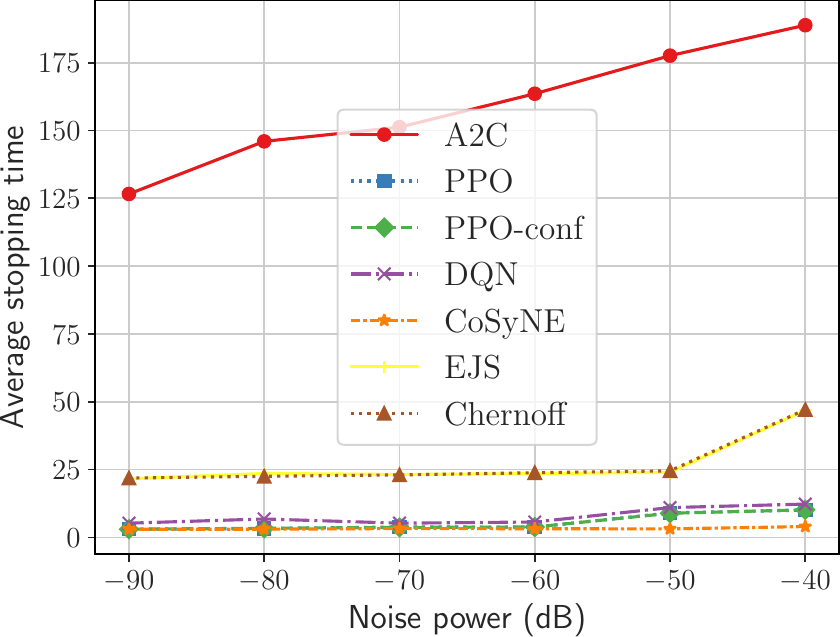}
        \caption{Average episode stopping time: $E=0.3$.}
    \end{subfigure}\hspace{1.25cm}
    \caption{\textcolor{black}{Similar metrics to Fig.~\ref{fig:single-binomial} but for the case of a Ricean fading channel model with \( \kappa^L = 5\,\mathrm{dB} \) and \( \kappa^E = -2\,\mathrm{dB} \).}}\label{fig:single-Ricean}
\end{figure*}

The results for $L=0.1$ and $E=0.3$ are shown in Fig. \ref{fig:single-Ricean}, where we varied the noise power from \(-90\,\mathrm{dB}\) to \(-40\,\mathrm{dB}\). It can be observed that our approach can satisfy all constraints, unlike the DRL baselines which fail to satisfy privacy requirements under strong noise conditions. It is also shown that our approach achieves vastly shorter stopping times than the DRL benchmarks. Interestingly, it can also terminate quicker than the classic heuristics that ignore the existence of the eavesdropping agent $E$. 

\paragraph{Strong-or-weak  radars for target detection}
We now consider a radar target detection application based on the strong-or-weak return model~\cite{chernoff-radar}, according to which the environment is either empty (i.e., no target is present), or exactly one of \( N_{\rm targ} \) possible targets exists inside it. The legitimate agent \( L \) has access to \( N_{\rm targ} \) distinct waveforms, each designed to be optimal for detecting a specific target. At each time instant \( t \), the agent \( L \) selects a waveform \( a_t \) for transmission and observes the reflected signal.

Under a Gaussian radar model, the received observation \( y_t \) follows a Gaussian distribution. If no target is present, \( y_t \sim \mathcal{N}(0, \sigma^L) \), i.e., zero-mean Gaussian with variance $\sigma^L$. If the transmitted waveform matches the true target \( \nu \), then \( y_t \sim \mathcal{N}(m_\nu^+, \sigma_L^2) \); otherwise, \( y_t \sim \mathcal{N}(m_\nu^-, \sigma_L^2) \), where \( m_\nu^+ \) and \( m_\nu^- \) represent the strong and weak signal means, respectively. The eavesdropping agent \( E \) observes the same waveform but through a degraded channel, resulting in a higher noise variance; this is denoted as \( \sigma_E^2 > \sigma_L^2 \).

\begin{figure*}
    \centering
    \begin{subfigure}{.33\textwidth}
        \centering
        \includegraphics[width=\textwidth]{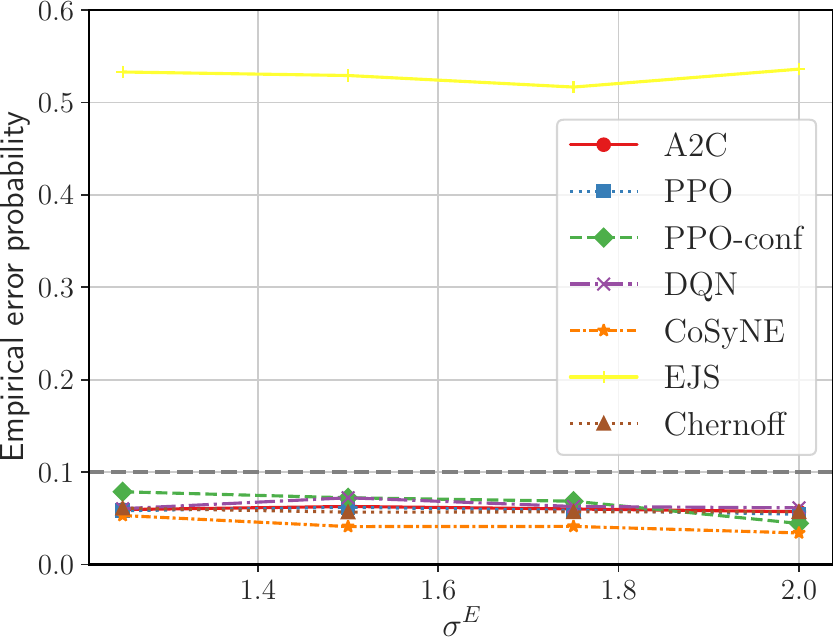}
        \caption{Legitimate error probability: $E=0.3$.}
    \end{subfigure}\hfill
    \begin{subfigure}{.33\textwidth}
        \centering
        \includegraphics[width=\textwidth]{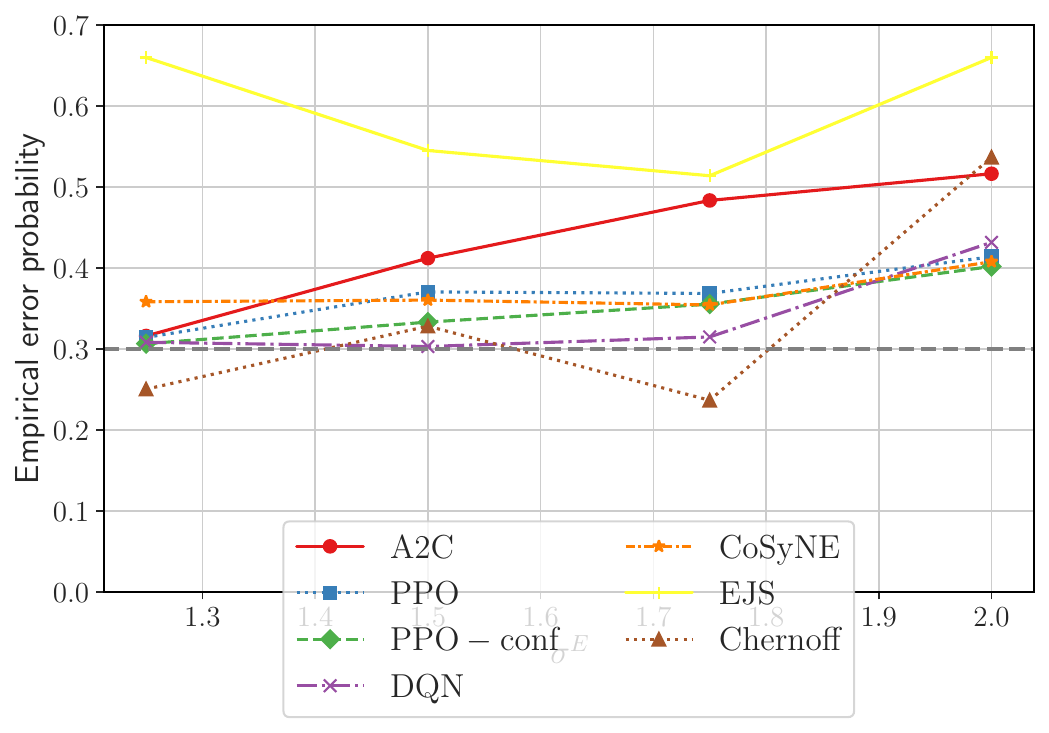}
        \caption{Eve's error probability: $E=0.3$.}
    \end{subfigure}\hfill
       \begin{subfigure}{.33\textwidth}
        \centering
        \includegraphics[width=\textwidth]{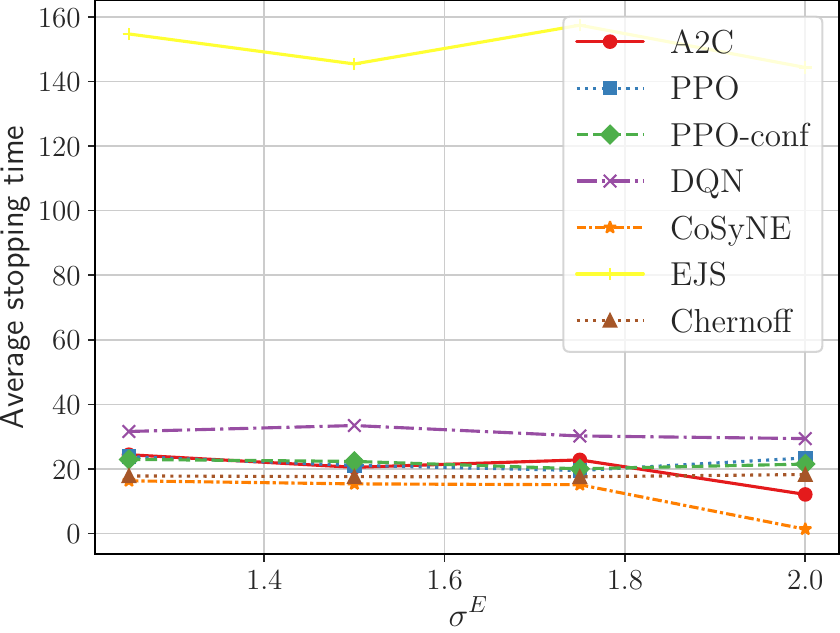}
        \caption{Average episode stopping time: $E=0.3$.}
    \end{subfigure}\hspace{1.25cm}
    \caption{\textcolor{black}{Similar metrics to Fig.~\ref{fig:single-binomial} but for the case of the strong-or-weak radar target detection application when the legitimate agent's noise standard deviation is set to \( \sigma_L^2 = 1 \).}}\label{fig:single-Rxdar}
\end{figure*}

In Fig.~\ref{fig:single-Rxdar}, we have set \( N_{\rm targ} = 5 \) and sampled $m_\nu^+$ uniformly in \([1, 2] \) and similarly for $m_\nu^-$ in \([0.1, 0.5] \) for each target \( \nu \). The legitimate agent's noise standard deviation was fixed at \( \sigma_L^2 = 1 \), while the eavesdropper's one varied from \( \sigma_E^2 = 1.25  \) to \( 2 \). The constraint thresholds were set to \( L = 0.1 \) and \( E = 0.3 \). It can be demonstrated that the proposed CoSyNE-based approach outperforms all benchmarks satisfying the constraints, while also achieving the shortest stopping times. Notably, the EJS method performs very poorly in this setting.

\color{black}
\subsection{Results for Decentralized EAHT}\label{subsec:DecResults}

\begin{figure}[!t]
    \centering
    \scalebox{0.42}{
    \includegraphics{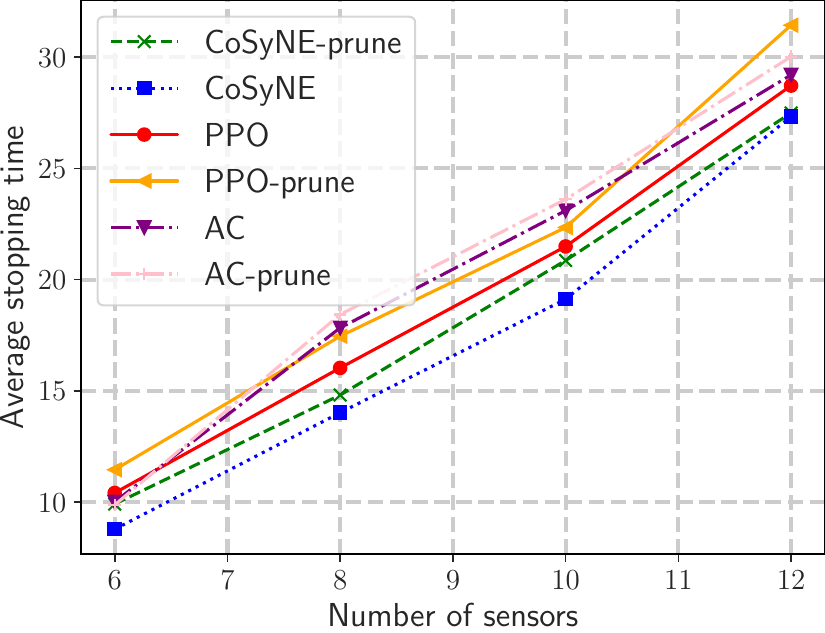}}
    \caption{Average episode stopping time of decentralized EAHT for the threshold values $L=0.1$ and $E=0.3$, as well as for different values of $S$, considering the binomial sensor model in~\eqref{yDef}.}  
    \label{fig:redDec1}
\end{figure}

\begin{figure}[!t]
    \centering
    \scalebox{0.42}{
    \includegraphics{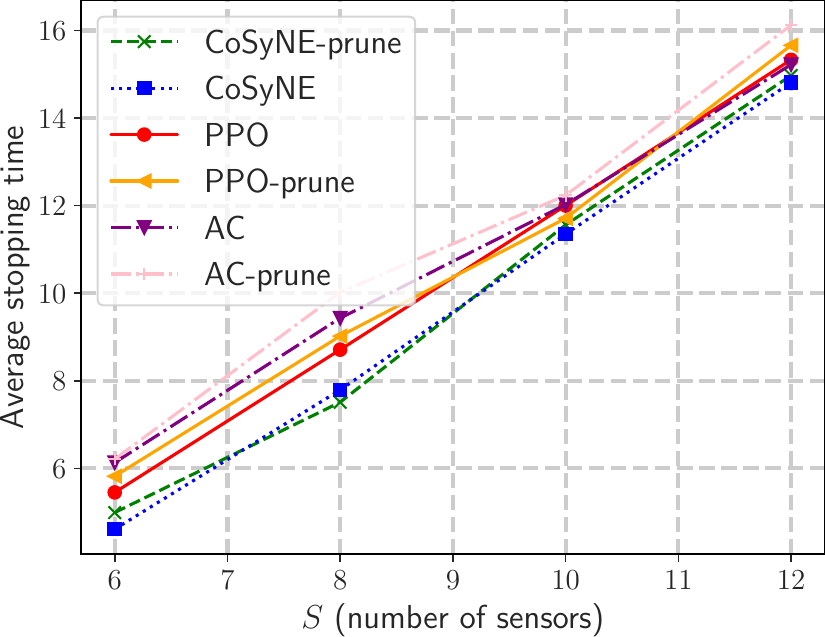}}
    \caption{Similar to Fig.~\ref{fig:redDec1}, but for \eqref{eq:GaussOBS}'s Gaussian sensor model. }  
    \label{fig:redDec2}
\end{figure}

In Fig.~\ref{fig:redDec1} and~\ref{fig:redDec2}, we have considered the same observation models with those used in the single-agent case as well as $K=4$ fully connected agents. We have also set the thresholds to $L=0.1$ and $E=0.3$, and varied the number of sensors $S$ from $6$ to $12$, yielding at most $2^{12}$ possible hypotheses in total. We have assumed that the first two agents have access to the first half of the sensors, and the other two have access to the rest of the sensors. Since both our NE-based methods and the DRL benchmark for multi-agent EAHT satisfied the accuracy and privacy constraints, we include only the average stopping time for the binomial and Gaussian sensor models in Fig.~\ref{fig:redDec1} and Fig.~\ref{fig:redDec2}, respectively. As shown, both the proposed unpruned and pruned (with only $10\%$ of the weights of the unpruned version) NE-based approaches achieve shorter stopping times than the designed DRL benchmarks. Interestingly, it is demonstrated that our pruned approach achieves better stopping times than the benchmarks, even if it had removed over $90\%$ of redundant weights in all simulated settings. 

\begin{figure}
    \centering
    \includegraphics[width=0.35\textwidth]{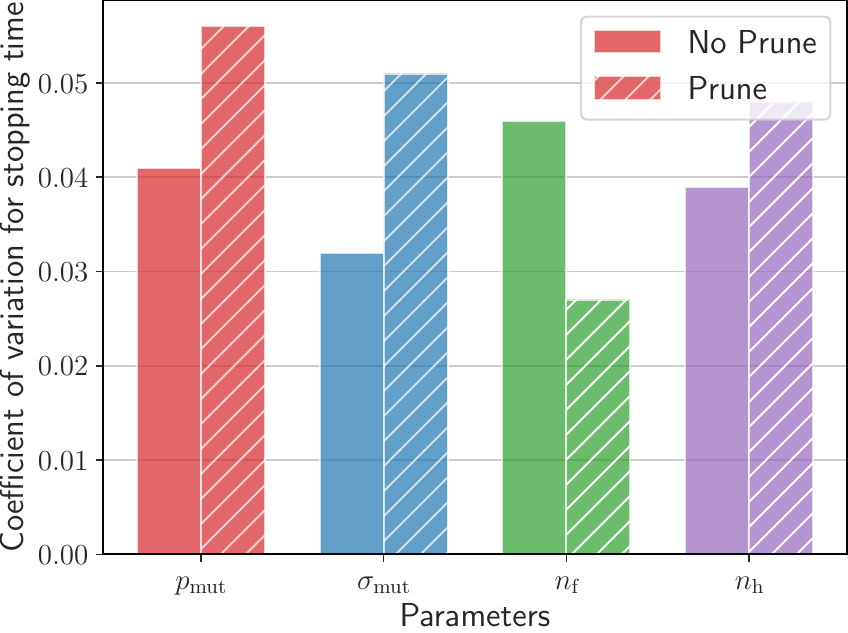}
    \caption{Coefficients of variation of the stopping time objective for both the unpruned and pruned NE-based EAHT schemes, considering $S=10$ sensors and hyperparameters' variation.}\label{fig:sensitivity-bars}
\end{figure}

To evaluate the robustness of the proposed NE-based EAHT schemes under variations in the hyperparameters, we have performed a detailed sensitivity analysis. This analysis was conducted on both the unpruned and pruned agents, using the binomial sensor model with $S=10$ sensors. For each hyperparameter, we varied its value while keeping all other parameters fixed. The training and testing procedures were repeated for each configuration, allowing us to assess the impact of individual hyperparameter changes on the system's performance. In particular, our sensitivity analysis included the following parameters:
\begin{enumerate}
    \item The probability $p_{\rm mut}$ was varied from $0.3$ to $0.6$.
    \item The standard deviation $\sigma_{\rm mut}$ was varied from $0.4$ to $0.8$.
    \item The hidden size of the extractor $n_{\rm f}$ was varied between $250$ and $400$.
    \item The hidden size of each individual branch $n_{\rm b}$ was varied between $250$ and $400$.
\end{enumerate}
Tables~\ref{tab:sensLegError} and~\ref{tab:sensEaveError} list respectively the maximum legitimate and minimum eavesdropping error probabilities for each of the latter four sensitivity studies. It is clearly demonstrated that both proposed NE-based EAHT schemes always satisfy the accuracy and privacy constraints. 
\begin{table}[]
    \centering
    \begin{tabular}{|c|c|c|c|c|}
    \hline
    Method & $p_{\rm mut}$ & $\sigma_{\rm mut}$& $n_{\rm f}$ & $n_{\rm b}$\\
\hline
\hline
         CoSyNE&  0.066&0.087&0.043&0.055 \\
         \hline  
         CoSyNE-prune&  0.072&0.061&0.046&0.044 \\
         \hline
    \end{tabular}
    \caption{The maximum (i.e., worst case) legitimate error probability for each sensitivity study. It can be seen that all values are below the threshold $L=0.1$.}
    \label{tab:sensLegError}
\end{table}
\begin{table}[]
    \centering
    \begin{tabular}{|c|c|c|c|c|}
    \hline
    Method & $p_{\rm mut}$ & $\sigma_{\rm mut}$& $n_{\rm f}$ & $n_{\rm b}$\\
\hline
\hline
         CoSyNE&  0.41&0.37&0.39&0.45 \\
         \hline  
         CoSyNE-prune&  0.52&0.41&0.33&0.34 \\
         \hline
    \end{tabular}
    \caption{The minimum (i.e., worst case) error probability at Eve for each sensitivity study. Clearly, all values are above the threshold $E=0.3$.}
    \label{tab:sensEaveError}
\end{table}
The CVs of the stopping time objective for each sensitivity study are illustrated in Fig.~\ref{fig:sensitivity-bars}, verifying the robustness of both proposed schemes. As shown, the coefficients are always smaller than $0.06$, implying that the stopping time objective is stable and robust to hyperparameter changes. All in all, the latter results underscore the stability and reliability of the proposed schemes, showcasing that their performance remains largely unaffected by hyperparameter variations. Such robustness is essential for practical implementations, ensuring consistent outcomes even under varying conditions.

\begin{figure}
    \centering
    \begin{subfigure}{.33\textwidth}
        \centering
        \includegraphics[width=\textwidth]{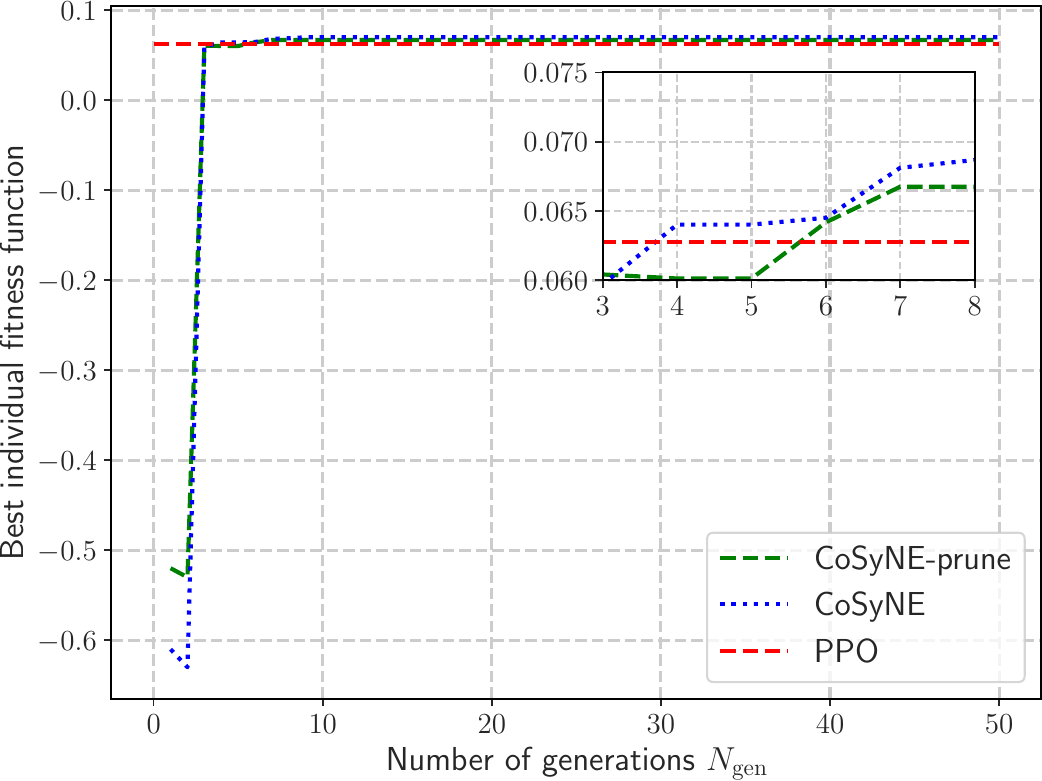}
        \caption{Fitness function.}\label{fitcurve}
    \end{subfigure}\hfill
    \begin{subfigure}{.33\textwidth}
        \centering
        \includegraphics[width=\textwidth]{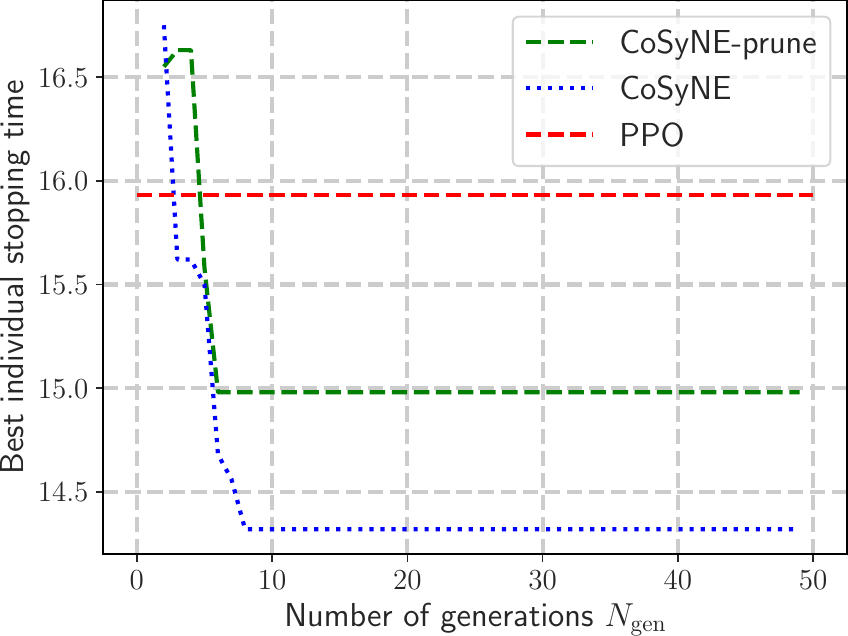}
        \caption{Stopping time.}\label{stcurve}
    \end{subfigure}
    \caption{Evolution of the EAHT performance versus the generation number $N_{\rm gen}$ for both the unpruned and pruned NE-based schemes, considering the same hyperparameter changes with Fig.~\ref{fig:sensitivity-bars} and $S=8$ sensors. Curves with our designed PPO-based scheme are also included for comparison purposes.}\label{fig:generation-curve}
\end{figure}

Finally, in Fig.~\ref{fig:generation-curve}, the evolution of the performance of our NE-based multi-agent EAHT schemes with respect to the generation number \(N_{\rm gen}\) is illustrated. In particular, Fig.~\ref{fitcurve} depicts the fitness functions of our schemes in comparison with the fitness of the trained PPO-based agents. It can be observed that, initially, the fitness scores are negative, which signifies a failure to satisfy the privacy constraint. However, within just two generations, both algorithms identify candidate policies that meet this requirement. At first glance, the differences between the fitness functions of the different schemes may appear insignificant, but this happens because the fitness equals to $1/\tau$, when the privacy constraint is satisfied. However, a small increase in fitness indicates a decrease of a few time steps in the detection delay, which can be extremely important in applications of abnormal activity detection. For enhanced visualization, Fig.~\ref{stcurve} presents the stopping times of all considered algorithms, starting from the third generation. As notably shown, within fewer than ten generations, both the pruned and unpruned agents surpass the performance of the multi-agent PPO-based benchmark. 

\color{black}
\subsubsection{Non-Stationary Graphs}\label{subsec:message_loss}
\begin{figure}
    \centering
    \begin{subfigure}{.33\textwidth}
        \centering
        \includegraphics[width=\textwidth]{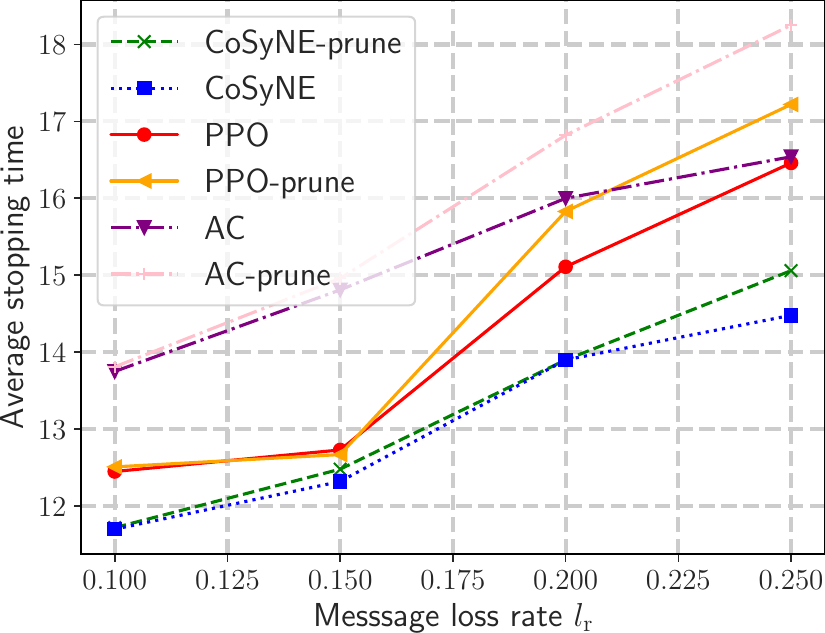}
        \caption{\textcolor{black}{Binomial sensor model.}}
    \end{subfigure}\hfill
    \begin{subfigure}{.33\textwidth}
        \centering
        \includegraphics[width=\textwidth]{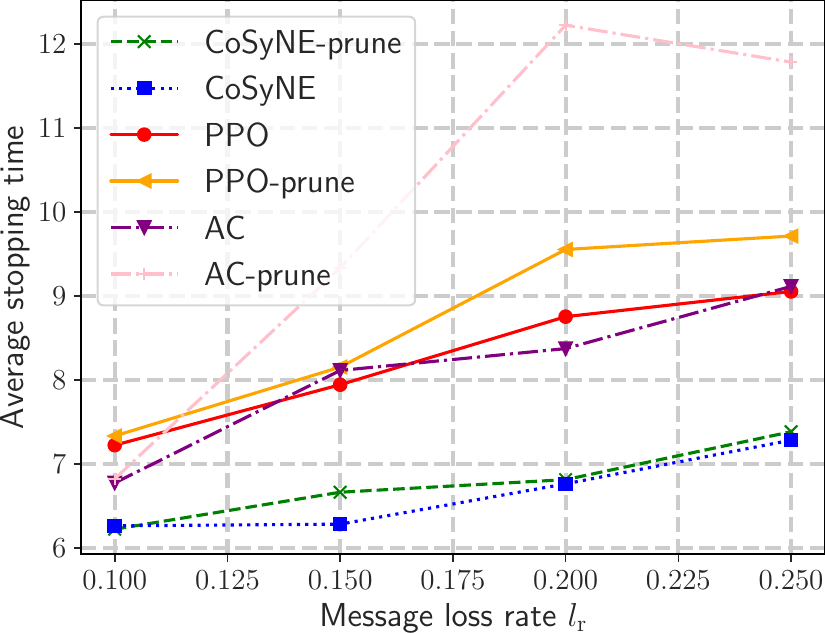}
        \caption{\textcolor{black}{Gaussian sensor model.}}
    \end{subfigure}
    \caption{\textcolor{black}{Average episode stopping time of decentralized EAHT for threshold values $L=0.1$, $E=0.3$, $S=6$, as well as for different values of the message loss rate $l_{\rm r}$.}}\label{fig:message losses}
\end{figure}

The previous experimentation considered only fully-connected agents. In this subsection, we investigate the generalization of the proposed NE-based decentralized agents when deployed in sparse time-varying communication graphs. To this end, we have assumed that each agent $k$ tries to broadcast the tuple $(a_t^k,y_t^k)$ to all other agents, but the message is lost with probability $l_{\rm r}$, implying that the agents have different information sets. Hence, the structure of the agent connection graph is different at each time instant $t$ due to the message losses. We have evaluated the proposed policies in Figs.~\ref{fig:redDec1} and~\ref{fig:redDec2} on new testing environments, considering the same observation models but varying the loss rate from $0.1$ to $0.25$. as depicted in Fig.~\ref{fig:message losses}, although all methods can satisfy the privacy constraint, the proposed NE methods achieve noticeably shorter stopping times.

\subsubsection{The Importance of Message Exchange}\label{subsec:message_results}
\begin{figure}
    \centering
    \includegraphics[width=0.75\linewidth]{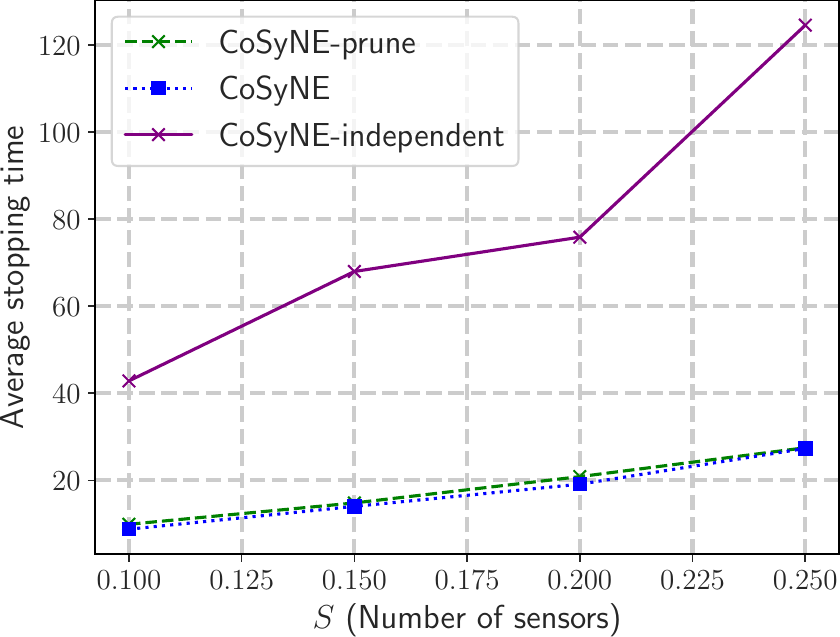}
    \caption{\textcolor{black}{Average stopping times for the decentralized EAHT with binomial observations considering networked versus independent CoSyNE agents. It is shown that lack of information exchange between agents more than doubles detection delay.}}
    \label{fig:abbl.}
\end{figure}

As previously noted, establishing secure and reliable communication channels between agents incurs costs. However, such communication can significantly enhance detection performance. To investigate the impact of message exchange, we have conducted a study on the binomial sensor model of~(\ref{yDef}) involving a fully independent group of agents utilizing our proposed DNN architecture evolved with CoSyNE. In this setup, all  agents had access to all sensors. Each agent independently probed one of the $S$ sensors and formed beliefs without message exchange. As illustrated in Fig. \ref{fig:abbl.}, this independence results in substantially increased stopping times. All other parameters (network structure, observation model, and algorithmic parameters) have been the same for both the independent and the networked CoSyNE agents.

\color{black}
\section{Conclusions and Future Work}
In this paper, we studied both single- and multi-agent EAHT problems and presented NE-based solutions. Specifically for the decentralized multi-agent problem, we devised a novel NE method for dealing with collaborative multi-agent tasks, which maintains all computational benefits of single-agent NE. We also extended it by providing a second algorithm that jointly optimizes the multi-agent policy and removes the unnecessary DNN parameters. The robustness and superiority of the proposed NE-based EAHT approaches over benchmarks was demonstrated through extensive numerical simulations. 

\textcolor{black}{While the EAHT problem was introduced in~\cite{evasive-active} with asymptotic bounds on the eavesdropper’s error exponent, no concrete strategies with provable performance guarantees have been developed so far. In contrast, classical AHT benefits from theoretically grounded policies such as the Chernoff test or EJS maximization. Our NE-based approach fills this gap offering a practical, flexible alternative that performs competitively across a wide range of settings, and generalizes well, despite lacking theoretical guarantees. Besides the aforementioned bounds, devising novel theoretical strategies with privacy guarantees is a very important area for future research, even if these strategies do not work as well as NE. }

One other research direction is to extend our experimental investigations and the theoretical analysis of~\cite{evasive-active} to other challenging active sensing tasks, like continuous high dimensional parameter estimation \cite{activeSamplingMultiSource}, change detection \cite{banditChangePoint}, and beam alignment \cite{activeLearningMMWave}. It is also worthwhile to examine scenarios with multiple heterogeneous and active eavesdroppers. \textcolor{black}{Finally, we intend to combine our data collection mechanism with active defense strategies~\cite{mtd1} in order to handle adaptive and progressively improving eavesdroppers. These eavesdroppers can gradually infiltrate the network and acquire more accurate observation models~\cite{TeneketzisDynamicDefence}}.

\bibliographystyle{ieeetr}  
\bibliography{references} 
\newpage
\begin{IEEEbiographynophoto}
{George Stamatelis } was born in Athens, Greece in the summer of 2000. He finished high school in $2018$ and  in $2022$ he got his BSc in computer science from the department of informatics and telecommunications of the National and Kapodistirian University of Athens with highest honours.
Since $2023$ he continuous his graduate studies in the same department. His research interests are machine learning, multiagent systems, anomaly detection and distributed signal processing.
\end{IEEEbiographynophoto}
\vskip 0pt plus -1fil
\begin{IEEEbiographynophoto}
{Angelos-Nikolaos Kanatas} studied electrical and computer engineering at the National and Technical University of Athens, and graduated with high honors.  His research interests include deep learning, audio and speech processing, natural language processing, and reinforcement learning. He has worked as a research associate with the Institute for Language and Speech Processing at the Athena Research Center.
\end{IEEEbiographynophoto}
\vskip 0pt plus -1fil
\begin{IEEEbiographynophoto}
{Ioannis Asprogerakas} studied electrical and computer engineering at the National and Technical University of Athens, where he is currently conducting his thesis on diffusion models. His research interests include generative models, computer vision, and multi-modal learning.
\end{IEEEbiographynophoto}
\vskip 0pt plus -1fil
\begin{IEEEbiographynophoto}
{George C. Alexandropoulos}~(Senior member, IEEE) received the Engineering Diploma (Integrated M.S.c), M.A.Sc., and Ph.D. degrees in Computer Engineering and Informatics from the School of Engineering, University of Patras, Greece in 2003, 2005, and 2010, respectively. He has held senior research positions at various Greek universities and research institutes, and he was a Senior Research Engineer and a Principal Researcher at the Mathematical and Algorithmic Sciences Lab, Paris Research Center, Huawei Technologies France, and at the Technology Innovation Institute, Abu Dhabi, United Arab Emirates, respectively. He is currently an Associate Professor with the Department of Informatics and Telecommunications, School of Sciences, National and Kapodistrian University of Athens (NKUA), Greece and with the Department of Electrical and Computer Engineering, University of Illinois Chicago, Chicago, IL, USA. His research interests span the general areas of algorithmic design and performance analysis for wireless networks with emphasis on multi-antenna transceiver hardware architectures, full duplex radios, active and passive Reconfigurable Intelligent Surfaces (RISs), Integrated Sensing And Communications (ISAC), millimeter wave and THz communications, as well as distributed machine learning algorithms.

\end{IEEEbiographynophoto}

\end{document}